\documentclass[twocolumn]{autart} 

\usepackage{subcaption}
\usepackage{algorithm}
\usepackage{algpseudocode}

\usepackage{lineno,hyperref}
\usepackage{amsmath,graphicx}
\usepackage{amssymb}
\usepackage{booktabs}
\usepackage{cite}
\usepackage{xcolor}
\usepackage{mathtools}
\usepackage{cases}
\usepackage{caption}
\newtheorem{theorem}{Theorem}[section]

\newtheorem{lemma}[theorem]{Lemma}

\newtheorem{assumption}[theorem]{Assumption}
\newtheorem{remark}[theorem]{Remark}

\newcommand\tldx{\tilde{\mathbf{x}}}

\newcommand\bc{\mathbf{c}}
\newcommand\bz{\mathbf{z}}
\newcommand\hbz{\widehat{\bz}}
\newcommand\bZ{\mathbf{Z}}
\newcommand\bX{\mathbf{X}}
\newcommand\bLam{\mathbf{\Lambda}}
\newcommand\obLam{\overline{\mathbf{\Lambda}}}

\newcommand\hbZ{\widehat{\mathbf{Z}}}
\newcommand\col{\operatorname{col}}

\newcommand\bE{\mathbb{E}}

\newcommand\bv{\boldsymbol{v}}
\newcommand\bx{\mathbf{x}}
\newcommand\ox{\overline{\bx}}
\newcommand\obx{\overline{\mathbf{X}}}
\newcommand\by{\mathbf{y}}

\newcommand\bu{\mathbf{u}}
\newcommand\teta{\tilde{\eta}}

\newcommand\onbf{\overline{\nabla \bf f}}

\newcommand\nbf{\nabla \mathbf{f}}

\newcommand\md{\mathcal{D}}
\newcommand\mB{\mathcal{B}}
\newcommand\bW{\mathbf{W}}

\newcommand{\blue}[1]{{\color{blue}#1}}
\definecolor{purple}{rgb}{0.6, 0.2, 0.8}

\usepackage{multibib}
\definecolor{UniBlue}{RGB}{83,121,170}
\definecolor{DarkGray}{RGB}{90,90,90}
\definecolor{LightGray}{RGB}{150,150,150}
\definecolor{oldTextGreen}{RGB}{115,155,15}
\definecolor{TextGreen}{RGB}{112,183,100}
\definecolor{oldOcean}{RGB}{23,142,189}
\definecolor{Ocean}{RGB}{30,106,181}
\definecolor{BG}{RGB}{215,215,215}
\definecolor{darkred}{RGB}{204,41,0}
\usepackage{hyperref}

\allowdisplaybreaks
\begin{document}
\begin{frontmatter}
\title{Non-convex composite federated learning\\ with heterogeneous data   \thanksref{footnoteinfo}} 

\thanks[footnoteinfo]{ This work was supported in part by the funding from Digital Futures and VR under the contract 2019-05319.  Corresponding author is Jiaojiao Zhang.  Parts of the material in this paper have been published at 2024 IEEE International Conference on Acoustics, Speech and Signal Processing. }

\author[a]{Jiaojiao Zhang}\ead{jiaoz@kth.se},    
\author[b]{Jiang Hu}\ead{hujiangopt@gmail.com},               
\author[a]{Mikael Johansson}\ead{mikaelj@kth.se}  

\address[a]{School of Electrical Engineering and Computer Science, KTH
Royal Institute of Technology}  
\address[b]{Massachusetts General Hospital and Harvard Medical School, Harvard University}             

\begin{abstract}
We propose an innovative algorithm for non-convex composite federated learning that decouples the proximal operator evaluation and the communication between server and clients.
Moreover, each client uses local updates to communicate less frequently with the server, sends only a single $d$-dimensional vector per communication round, and overcomes issues with {client drift}.
In the analysis,  challenges arise from the use of decoupling strategies and local updates in the algorithm, as well as from the non-convex and non-smooth nature of the problem. We
establish sublinear and linear convergence {to a bounded residual error} under general non-convexity and the proximal Polyak-{\L}ojasiewicz inequality, respectively. 
In the numerical experiments, we demonstrate the superiority of our algorithm over state-of-the-art methods on both synthetic and real datasets.
\end{abstract}
\begin{keyword}                           
non-convex composite federated learning;  heterogeneous data; local update.               
\end{keyword}   
\end{frontmatter}

\section{Introduction}
Federated Learning (FL) is a widely used machine learning framework in which a central server coordinates numerous clients to collaboratively train a model without sharing their local data~\cite{li2020federated-fedprox}. 
The appeal of FL lies in its distributed computations and potential for privacy protection, making it attractive in various applications. Notable examples include enhancing the efficiency and resilience of smart grids~\cite{badr2023privacy}, optimizing data acquisition and processing in wireless sensor networks~\cite{hong2022distributed}, and reinforcing privacy safeguards in data-sensitive environments~\cite{wang2023decentralized}.

However, compared to conventional distributed learning, FL encounters significant challenges, notably a communication bottleneck at the server and a sensitivity to data heterogeneity among clients \cite{li2020federated-fedprox,zhang2023fedaudio}.
In an effort to improve communication efficiency, McMahan et al. proposed Federated Averaging (FedAvg)  \cite{mcmahan2017communication}, where clients execute multiple local updates before transmitting their updated states to the server. 
In cases where the clients possess similar data sets, implementing local updates is a practical method to enhance communication efficiency~\cite{stich2018local}. 
{
In heterogeneous settings, where data distributions vary across clients, many FL algorithms suffer from \emph{client drift}~\cite{li2019convergence,karimireddy2020scaffold}. This phenomenon occurs when individual clients, by performing multiple local updates, move excessively toward minimizing their local loss. This overfitting at the local level causes the global model, which is the average of these local models, to deviate from the optimal solution. A pedagogical illustration of client drift can be found in~\cite[Figure 2]{karimireddy2020scaffold}.
}
To overcome this issue, several solutions have been proposed in the literature  \cite{li2020federated-fedprox,karimireddy2020scaffold,pathak2020fedsplit,karimireddy2020mime}.

The majority of existing FL algorithms focus on smooth problems.
However, real-world applications frequently demand the optimization of non-smooth loss functions, especially when we want to find solutions with specific properties such as sparsity or low-rank \cite{fosson2020sparse,yuan2021federated}. 
This motives us to consider composite FL problems on the form 
\begin{equation}\label{eqn:basic_opt}
\begin{aligned}
\operatorname*{minimize}_{\bx\in\mathbb{R}^d} \; &F(\bx):=f(\bx)+g(\bx). 
\end{aligned}
\end{equation}
Here, $\bx\in \mathbb{R}^d$ represents the decision vector, i.e., the model parameters in a machine learning application, 
$f(\bx):= \frac{1}{n}\sum_{i=1}^nf_i(\bx)$ is the average loss across the $n$ clients that is assumed to be smooth but non-convex, and $g$ is a convex but possibly non-smooth regularizer. 
To make the data dependence explicit, we let 
$\md_i = \bigcup_{l=1}^{m_i} \md_{il}$, where  $\md_{i1}, \dots, \md_{im_i}$ denote the $m_i$  data points of client $i$,  $f_{il} (\bx;\md_{il})$ denote the sample loss of client $i$ associated with the data point $\md_{il}$, and
$f_i(\bx):=\tfrac{1}{m_i} \sum_{\md_{il}\in \md_i} f_{il} (\bx;\md_{il})$. 
Notably, no assumptions are made on the similarity among the data sets $\md_i$.

Federated learning with non-smooth, non-convex loss functions and heterogeneous data is challenging from both an algorithm design and a convergence analysis perspective.
We can glimpse into these challenges through an approach called Federated Mirror Descent (FedMid)~\cite{yuan2021federated}, which extends the FedAvg by replacing the local stochastic gradient descent (SGD) steps in the local update with proximal SGD~\cite{beck2009fast, yousefian2012stochastic}. 
FedMid encounters an issue that the authors call ``curse of primal averaging''~\cite{yuan2021federated}. 
To illustrate this phenomenon,  consider a setup-{\bf u}p where $g$ is the $\ell_1$-norm. Even if each client generates a sparse local model after the local updates,  averaging these local models at the server usually leads to a solution that is no longer sparse. 
Similar to FedAvg, FedMid also encounters the client drift issue. These limitations significantly impair the practical performance of FedMid.
%
In the analysis of FedMid, a key difficulty appears due to 
the coupling between the proximal operator and communication. If the server averages local models that have been updated using proximal operators during the local updates, it cannot recover the average gradients across all the clients due to the nonlinearity of the proximal operator mapping. This coupling, combined with the non-convex and non-smooth nature of the problem~\eqref{eqn:basic_opt}, makes the analysis challenging.

\subsection{Related work}

We broadly categorize the related work for solving \eqref{eqn:basic_opt} into algorithms that consider smooth ($g=0$) and non-smooth ($g\neq 0$) losses.

{\bf  Smooth FL problems. }
FedAvg was originally proposed in \cite{mcmahan2017communication} for federated
learning of deep networks, but without convergence analysis. A comprehensive analysis of FedAvg for strongly convex problems was conducted in \cite{stich2018local} under the assumption of homogeneous data distribution among clients. %
However, when data is heterogeneous, the use of local updates in FedAvg leads to the issue of client drift. As highlighted in~\cite{zhang2021fedpd}, without strong assumptions on the problem structure, the behavior of FedAvg can become highly erratic due to the impact of client drift. 
The impact of client drift on FedAvg was theoretically analyzed in~\cite{li2019convergence,zhang2021fedpd} under the assumption of bounded data heterogeneity. While \cite{li2019convergence} focused on strongly convex problems,  \cite{zhang2021fedpd} explored non-convex problems. 
Another line of work attempts to reduce or eliminate client drift by modifying FedAvg at the algorithmic level~\cite{li2020federated-fedprox,c1, karimireddy2020scaffold,pathak2020fedsplit,karimireddy2020mime}.  
{For example, \cite{li2020federated-fedprox} and \cite{c1} penalized the difference between the local models on clients and the global model on the server to ensure that the local models remain close, thereby reducing client drift. Both \cite{li2020federated-fedprox} and \cite{c1} established convergence for non-convex problems.} Scaffold \cite{karimireddy2020scaffold} and Mime \cite{karimireddy2020mime} tackle client drift by designing control variates to correct the local direction during the local updates, achieving convergence for non-convex problems.  
A drawback of these approaches is their need to communicate also the control variates, which increases the overall communication cost.  
In contrast, Fedsplit \cite{pathak2020fedsplit} adopts Peaceman-Rachford splitting \cite{he2014strictly} to address the client drift through a consensus reformulation of the original problem, exchanging only one local model per communication round for convex problems.
None of the methods discussed above handles composite FL problems.

{\bf Composite FL problems.} 
Compared to the extensive research on smooth FL problems,  composite problems have received significantly less attention. An effort to bridge this gap is the Federated Dual Averaging (FedDA) introduced in~\cite{yuan2021federated}. In FedDA, each client employs dual averaging~\cite{nesterov2009primal} during the local updates,  while the server averages the local models in the dual space and applies the proximal step. 
Convergence was established for convex problems, with the ability to handle general loss functions as long as the gradients are bounded. However, under data heterogeneity, the convergence analysis is limited to quadratic loss functions.
The Fast Federated Dual Averaging (Fast-FedDA) algorithm~\cite{bao2022fast} incorporates weighted summation of past gradient and model information during the local updates, but it introduces additional communication overhead.  While Fast-FedDA achieves convergence for general loss functions, it still relies on the assumption of bounded heterogeneity and can only handle strongly convex problems.
Federated Douglas-Rachford (FedDR) was introduced in \cite{tran2021feddr} and is able to avoid the bounded heterogeneity assumption. A subsequent development, FedADMM \cite{wang2022fedadmm}, utilizes FedDR to solve the dual problem of \eqref{eqn:basic_opt} and was demonstrated to have identical performance to FedDR. 
For both FedDR and FedADMM, convergence is established for non-convex problems and the local updates implement an inexact evaluation of the proximal operator of the smooth loss with adaptive accuracy. However, to ensure convergence, the accuracy needs to increase by every iteration, resulting in an impractically large number of local updates.

{The paper~\cite{c3} utilizes the mini-batch stochastic proximal point (MSPP) method as the local
stochastic optimal oracle for FedProx~\cite{li2020federated-fedprox}. The proximal term is used to mitigate client drift, but to guarantee convergence, the local subproblem needs to be solved exactly, which implies a large number of local updates. The paper~\cite{c3} established convergence 
for Lipschitz-continuous and weakly convex loss functions.
The work~\cite{c2} investigated differential privacy in non-convex federated learning. It proposed two variance-reduced algorithms for solving composite problems under the proximal Polyak-{\L}ojasiewicz (PL) inequality and general non-convexity, respectively. However, the algorithms in~\cite{c2} utilize a single local update, leading to frequent communication between clients and the server, and
assume that the smooth part of the composite loss is Lipschitz continuous. }

\subsection{Contributions} 
We propose an efficient algorithm with an innovative decoupling of the proximal operator evaluation and communication. 
This decoupling is achieved by 
letting each client manipulate two local models, before and after applying the proximal mapping, and sending the pre-proximal local model to the server. Moreover, each client uses local updates to communicate less frequently with the server, sends only a single $d$-dimensional vector per communication round, and overcomes the issue of client drift by an appropriately designed correction term. 
In our analysis, we address challenges arising from decoupling strategies, local updates, and the non-convex, non-smooth nature of the problem. Using
a carefully crafted {auxiliary} function, we establish sublinear and linear convergence {to a residual error} when the problem is generally non-convex and satisfies the proximal PL inequality, respectively. 
The convergence bounds explicitly quantify the impact of problem and algorithm parameters on the convergence rate and accuracy.
We verify the superiority of our algorithm over state-of-the-art methods on synthetic and real datasets in numerical experiments.

{\bf Notation.} 
We let $\|\cdot\|$ be  $\ell_2$-norm and $\|\cdot\|_1$ be $\ell_1$-norm. For positive integers $d$ 
and $n$, we let $\mathbf{I}_d$  be the $d\times d$ identity matrix, $\mathbf{1}_n$ ($\mathbf{0}_n$) 
be the all-one (all-zero) $n$-dimensional column vector,
and $[n] = \{1,\ldots, n\}$. We use $\otimes$ to denote the Kronecker product and let $\bW=\frac{1}{n}\mathbf{1}_n \mathbf{1}_n^T\otimes \mathbf{I}_d $.  For a set   $\mB$, we use $|\mB|$ to denote the cardinality.  For a  convex function $g$, we use  $\partial g$ to denote the subdifferential. For a random variable $\bv$,  we use $\mathbb{E}[\bv]$ to denote the expectation and $\mathbb{E}[\bv|\mathcal{F}]$ to denote the expectation given event $\mathcal{F}$. For vectors $\bx_1,\ldots,\bx_n\in\mathbb{R}^d$, we let  $\bX=\col\{\bx_i\}_{i=1}^{n}=[\bx_1;\ldots;\bx_n]\in\mathbb{R}^{nd}$. Specifically, for a vector $\ox\in \mathbb{R}^d$, we let  $\obx=\col\{\ox\}_{i=1}^{n}=[\ox;\ldots;\ox] \in \mathbb{R}^{nd}$. 
For gradients $\nabla f_1(\bx_1),\ldots, \nabla f_n(\bx_n) \in \mathbb{R}^d$, we let $\nabla{\bf f}({\bf X})=\col\{ \nabla f_i(\bx_i) \}_{i=1}^n$ and  $\onbf({\bf X})=\col\{ \frac{1}{n}\sum_{i=1}^n \nabla f_i(\bx_i) \}_{i=1}^n$.
For a vector $\boldsymbol{w}$ and a positive scalar $\teta$, we let  $P_{\teta}(\boldsymbol{w})=\operatorname{argmin}_{\mathbf{u}\in \mathbb{R}^d} \teta g(\mathbf{u})+\frac{1}{2}\|\boldsymbol{w}-\mathbf{u} \|^2 $. Specifically, for $\col\{ \boldsymbol{w}_i\}_{i=1}^n$,  we let $P_{\teta}(\col\{\boldsymbol{w}_i\}_{i=1}^n)=\col\{ P_{\teta}(\boldsymbol{w}_i)\}_{i=1}^n$. 

\section{Proposed algorithm}

In this section, we describe the implementation of our algorithm from the perspective of a single client. We then highlight the ideas behind the algorithm design through an equivalent compact form of our algorithm. 

\subsection{Per-client implementation of proposed algorithm}
In general, our algorithm comprises communication rounds indexed by $r$ and local updates indexed by $t$. 
During each round $r$, clients execute $\tau$ local update steps before updating the server. 
Each client $i$ maintains two models during the local updates, representing the local model state before and after the application of the proximal mapping. We call these models pre-proximal, denoted $\hbz_{i,t}^r$, and post-proximal, denoted $\bz_{i,t}^r$. {
Local mini-batch stochastic gradients of size $b$ are computed at the post-proximal local model $\bz_{i, t}^r$. A client-drift correction term $\bc_i^r$ is then added to the gradients, as shown on Line 9 in Algorithm 1, to correct the update direction for the pre-proximal local model $\hbz_{i,t}^r$.
}
{ Although during the current local updates, client $i$ does not have access to the current gradient information of other clients, we can obtain the old gradient information from other clients in the previous communication round and use it to correct the current local gradient direction. Our correction term $\bc_i^r$ is based on this idea. It can be locally constructed using the global model broadcasted by the server, as the global model contains the average gradient information due to the previous communication.}
At the end of the round, the final pre-proximal model, $\hbz_{i,\tau}^r$, is transmitted to the server.  

At the $r$-th communication round, the server also manipulates two models: a pre-proximal global model $\ox^r$ and a post-proximal global model $P_{\teta}(\ox^r)$. 
The server averages the pre-proximal local models $\hbz_{i,\tau}^r$ and utilizes the average information to update the post-proximal global model $P_{\teta}(\ox^r)$,  ensuring that the server-side algorithm behaves similarly to a centralized proximal SGD method.  Finally, the server broadcasts the pre-proximal global model $\ox^{r+1}$ to all clients that use it to update their correction terms $\bc_i^{r+1}$. 

\begin{figure}[htbp]
\centering
\includegraphics[width=8.7cm]{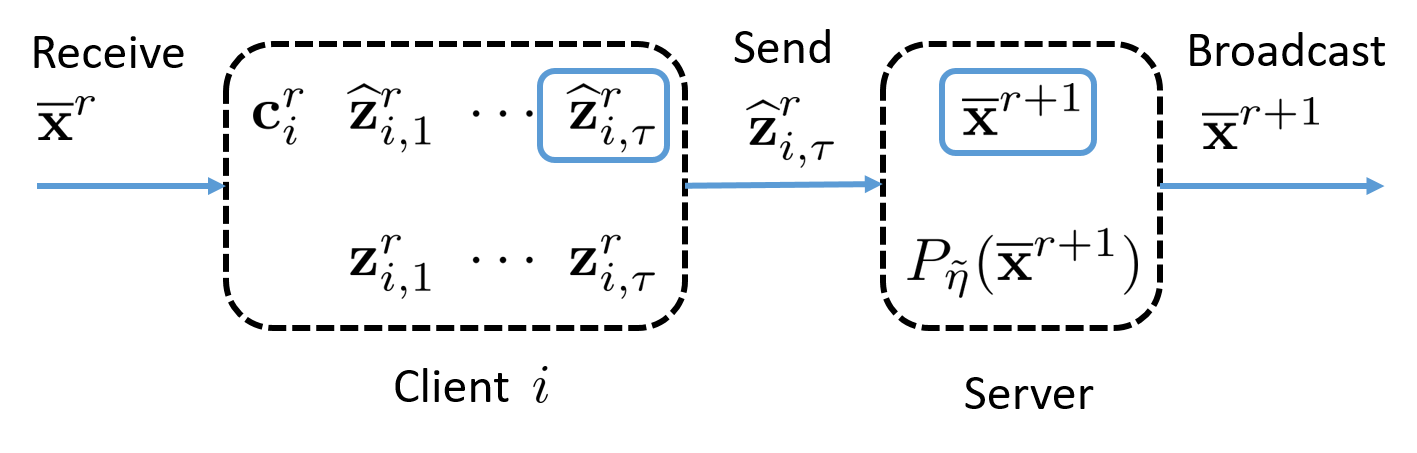}
\caption{{ Block diagram of Algorithm~\ref{alg-fl}.}}
\label{auto-diagram}
\end{figure}

{The per-client implementation of the algorithm is outlined in Algorithm \ref{alg-fl}.  Fig.~\ref{auto-diagram} illustrates how the pre-proximal local models $\hbz_{i,\tau}^r$ are sent by each client $i$ to the server and the pre-proximal global model $ \overline{\bx}^{r+1}$ is broadcasted by the server to all clients.}
\begin{algorithm}[h]
\caption{Proposed algorithm}
\label{alg-fl}
\begin{algorithmic}[1]
\State $ \textbf{Input:}$ $R$, $ \tau$, $\eta$, $\eta_g$, $\ox^{1}$, and $\bc_i^1=\mathbf{0}_d$ for all $i\in[n]$
\State Set $\teta=\eta \eta_g \tau$
\For {$r = 1, 2, \ldots, R$ }
\State {\bf Client $i$}
\State Set $\hbz_{i, 0}^r=P_{\teta}(\ox^r)$ and $\bz_{i,0}^r=P_{\teta}(\ox^r)$
\For {$ t= 0, 1, \ldots, \tau-1$ }
\State Sample a subset data $\mB_{i,t}^r \subseteq \md_i $ with $|\mB_{i,t}^r |=b$ 
\State Update $\!\nabla \!f_i(\bz_{i,t}^r;\!\mB_{i,t}^r)\!=\!\frac{1}{b}\! \sum\limits_{\md_{il}\in \mB_{i,t}^r}\!\nabla f_{il}(\bz_{i,t}^r;\!\md_{il})$
\State Update 
$
\begin{aligned}
&\hbz_{i,t+1}^r= 
\hbz_{i,t}^r-\eta \left(\nabla f_i(\bz_{i,t}^r;\mB_{i,t}^r)+\bc_i^r\right)
\end{aligned}
$
\State Update $\bz_{i,t+1}^r=P_{(t+1)\eta}{\left(\hbz_{i,t+1}^r\right)}$
\EndFor
\State Send $ \hbz_{i,\tau}^r$ to the server
\State {\bf Server}
\State Update $\ox^{r+1}\!=\!P_{\teta}(\ox^r)+\eta_g\!\left(\tfrac{1}{n}\! \sum_{i=1}^{n} \hbz_{i,\tau}^r     \!-\!P_{\teta}(\ox^r)\right)$ 
\State Broadcast $\ox^{r+1}$  to all the clients	
\State {\bf Client $i$}
\State Receive $\ox^{r+1}$ from the server
\State Update $\bc_i^{r+1}=\frac{1}{\eta_g\eta\tau}( P_{\teta}(\ox^{r})-\ox^{r+1})-\frac{1}{\tau} \sum_{t=0}^{\tau-1}\nabla f_i(\bz_{i,t}^{r};\mB_{i,t}^{r})$
\EndFor
\State \textbf{Output:} $P_{\teta}(\ox^{R+1})$
\end{algorithmic}
\end{algorithm}

\subsection{The intuition behind the algorithm design} \label{sec:intuition}
{As shown in Appendix~\ref{app:prf-illustration}, the iterates generated by Algorithm~\ref{alg-fl} can be described in compact form}
\begin{equation}\label{eqn:illustration}
\hspace{-2mm}
\left\{
\begin{aligned}
\hbZ_{t+1}^r &\!=\!\hbZ_t^r\!-\!\eta \Big(\nabla \mathbf{f}\left(\bZ_t^r; \mB_{t}^r\right)\!+\!\frac{1}{\tau} \sum_{t=0}^{\tau-1} \onbf\left(\bZ_t^{{r-1}} ; \mB_{t}^{r-1} \right)\\
& \quad- \frac{1}{\tau} \sum_{t=0}^{\tau-1} \nabla \mathbf{f} \left(\bZ_t^{{r-1}} ; \mB_{t}^{r-1}\right)  \Big),  \hspace{0.5cm}\forall t\in[\tau]-1, \\
\bZ^r_{t+1}&=P_{(t+1)\eta}\left(\hbZ^{r}_{t+1} \right), \hspace{2.1cm}~\forall t\in[\tau]-1,\\
\obx^{r+1} &= P_{\teta} (\obx^{r})-{\eta_g}\eta \sum_{t=0}^{\tau-1} \overline{\nabla \mathbf{f}}\left(\bZ_t^r ; \mB_{t}^r\right), 
\end{aligned}
\right.
\end{equation}
where  $\bZ_t^r=\col\{\bz_{i,t}^r\}_{i=1}^n$, $\hbZ_t^r=\col\{ \hbz_{i,t}^r\}_{i=1}^n$, $\obx^r=\col\{\ox\}_{i=1}^{n}$,  
$\nabla \mathbf{f}(\bZ_t^r;\mB_{t}^r) =\col\{\nabla f_{i}(\bz_{i,t}^r;\mB_{i,t}^r)\}_{i=1}^n$,  and  $ \onbf(\bZ_t^r;\mB_{t}^r)$ $=\col\left\{ \frac{1}{n} {\sum_{i=1}^{n}\!\nabla f_{i}(\bz_{i,t}^r;\mB_{i,t}^r)}\right\}_{i=1}^n $. At $r=1$, we set $\nabla f_i \left(\bz_{i,t}^{{0}}; \mB_{i,t}^{0}\right) =\mathbf{0}_d$ for all $t\in [\tau]-1$ so that $\frac{1}{\tau} \sum_{t=0}^{\tau-1} \onbf\left(\bZ_t^{0} ; \mB_{t}^0 \right)- \frac{1}{\tau} \sum_{t=0}^{\tau-1} \nabla \mathbf{f} \left(\bZ_t^{0} ; \mB_{t}^{0}\right)=\mathbf{0}_{nd}$.

After the last step in \eqref{eqn:illustration}, the server update satisfies
\begin{equation}\label{eq:bar-27}
\begin{aligned}
{P_{\teta}(\ox^{r+1})}\!=\!P_{\teta}\Big( P_{\teta}(\ox^r)\!-\!\teta \underbrace{\frac{1}{n\tau} \sum_{t=0}^{\tau-1} \sum_{i=1}^n \nabla f_i\left(\bz_{i,t}^r;\mB_{i,t}^r\right)}_{:=\bv^r}\!\Big).
\end{aligned}
\end{equation}	
For the sake of clarity, we define a virtual iterate $\tilde{\bx}^{r+1}$ that follows the centralized  proximal gradient descent (PGD) update
\begin{equation}\label{eq:pgd}
\begin{aligned}
\tilde{\bx}^{r+1}
: =P_{\teta} \left( P_{\teta}(\ox^r)-\teta \nabla f(P_{\teta}(\ox^r))\right).
\end{aligned}
\end{equation}
The server update \eqref{eq:bar-27} is similar to the centralized PGD iterate \eqref{eq:pgd}, except that the direction $\bv^r$ in our algorithm uses stochastic gradients and is the average across local updates and clients. The server broadcasts the pre-proximal global model $\ox^r$ since the clients need it to construct the correction term  (Line 18 in Algorithm \ref{alg-fl}).

With the algorithm re-written in this compact form, we are now ready to explain its key features.

{\bf {1)} Decoupling prox evaluation and communication.}
To achieve decoupling, each client $i$ manipulates a pre-proximal local model $\hbz_{i,t}^r$ during the $\tau$ local updates and then transmits $\hbz_{i,\tau}^r$  to the server.
In the update of $\hbz_{i,t}^r$ (the first step of \eqref{eqn:illustration}), each client $i$ evaluates the mini-batch SGD at the post-proximal local model $\bz_{i,t}^r$ and adds the correction term.  Since the average of the correction terms across clients is always zero, the server can obtain the average gradient $\sum_{t=0}^{\tau-1}\frac{1}{n} {\sum_{i=1}^{n} \nabla f_{i}(\bz_{i,t}^r;\mB_{i,t}^r)}$ at the post-proximal models $\{\bz_{i,t}^r\}_i$ by simply averaging $\hbz_{i,\tau}^r$. In this way, our algorithm is able to convey the average gradient to the server without any distortion.

{\color{black}
In contrast, if each client  $i$ would naively use proximal SGD with client-drift correction during the local updates, i.e.,   
\begin{equation*}
\begin{aligned}
\bz_{i,t+1}^r
&= P_{(t+1)\eta} \Big(\bz_{i,t}^r-\eta (\nabla f_i(\bz_{i,t}^r;\mB_{i,t}^r)\\& + \frac{1}{\eta_g\eta\tau}( P_{\teta}(\ox^{r-1})-\ox^{r}  )-\frac{1}{\tau} \sum_{t=0}^{\tau-1} \nabla f_i(\bz_{i,t}^{r-1};\mB_{i,t}^{r-1}) ) \Big)
\end{aligned}
\end{equation*}
and transmit $ \bz_{i,\tau}^r$ to the server after $\tau$ local updates,  then the server could no longer extract the average gradient due to the nonlinearity of the proximal operator.  }

It should be noted that the mismatch in the first step of \eqref{eqn:illustration}, i.e., updating $\hbz_{i,t}^r$ with the gradients evaluated at $\bz_{i,t}^r$, introduces challenges in the analysis. Nevertheless, by the use of novel analytical techniques, we manage to address this issue in Lemma \ref{lem:phi-bar-x}. 

{\bf { 2)} Correcting for client drift.} 
The re-formulation (\ref{eqn:illustration}) exposes how the correction terms affect the iterates. Effectively, each client $i$  introduces gradient information from other clients via
$\tfrac{1}{n \tau} \sum_{t=0}^{\tau-1} \sum_{i=1}^{n}\nabla f_i(\bz_{i,t}^{r-1};\mB_{i,t}^{r-1})$
and then replaces its own previous local contribution $\tfrac{1}{\tau} \sum_{t=0}^{\tau-1} \nabla f_i \left(\bz_{i,t}^{r-1}; \mB_{i,t}^{r-1}\right)$ with the new $\nabla f_i(\bz_{i,t}^{r};\mB_{i,t}^{r})$. { Intuitively,  each local update uses global gradient information to obtain the post-proximal local model $\bz_{i,t+1}^r$, which is a single iterate aimed at minimizing the global loss function \(\frac{1}{n}\sum_{i=1}^n f_i+g\), rather than \(f_i+g\). This allows our algorithm to eliminate the client drift.}

{\bf {3)} Reduced signalling.}
Unlike Scaffold \cite{karimireddy2020scaffold} and Mime \cite{karimireddy2020mime}, who overcome client drift for smooth problems at the cost of additional communication for control variates, our algorithm only exchanges one $d$-dimensional vector per communication round and client. 
This is made possible by a clever algorithm design that allows each client to reconstruct its own correction term from the broadcasted $\ox^r$ without additional communication.

{\bf {4)} Novel parameter selection during local updates.}
In \eqref{eqn:illustration},  the updates of the post-proximal local models $\bZ_{t+1}^r$ use the parameter $(t+1)\eta$ for computing the proximal operator $P_{(t+1)\eta}{(\hbZ_{t+1}^r)}$.
{
The main idea of using $(t+1)\eta$ is to make $\bz_{i,t}^r$ evolve similarly as a centralized proximal gradient descent (PGD) method applied to the global model $P_{\teta}(\overline{\bx}^r)$:
\begin{equation}\label{eq-rev-2}
P_{\hat{\eta}}\Big(P_{\teta}(\overline{\bx}^r)-\hat{\eta}\nabla f(P_{\teta}(\overline{\bx}^r))\Big).   
\end{equation}
Note that the same step size $\hat{\eta}>0$ is used both in the proximal operator and in front of the gradient.  Since we perform $(t+1)$ steps of local updates with step size $\eta$ on $\hat{\bz}^r_{i,0}$ to obtain $\hat{\bz}_{i,t+1}^r$, the parameter $\hat{\eta}$ for proximal operator should be set to $(t+1)\eta$ to match (\ref{eq-rev-2}).

Specifically, by repeated use of Line 9 in Algorithm 1, 
\begin{align}\label{eq-review1}
\hbz^r_{i,t+1} & = \hbz^r_{i,0} - \eta \left( \sum_{\ell=0}^{t} \left(\nabla f_i\left(\bz^r_{i,\ell}; \mB_{i,\ell}^r\right) + \bc_i^r\right) \right) \\
& = \hbz^r_{i,0} \!-\! (t+1)\eta \Big( 
\frac{1}{t+1} \sum_{\ell=0}^{t} \big(\underbrace{\nabla f_i\left(\bz^r_{i,\ell}; \mB_{i,\ell}^r\right) + \bc_i^r }_{\approx\nabla f(\bz^r_{i,\ell})} \big) \Big),\nonumber
\end{align}
where we divide and multiply $(t+1)$ in the second equality of \eqref{eq-review1}. 
If we disregard that we use stochastic gradients and evaluate the gradient at different iterates $\bz_{i,0}^r, \ldots, \bz_{i,t}^r$,  $\nabla f_i(\bz_{i,\ell}^r; \mathcal{B}_{i,\ell}^r)$ in the second equality in \eqref{eq-review1} can be replaced by  $\nabla f_i(\bz_{i,0}^r)$ leading to
\begin{equation}
\begin{aligned}
\hat{\bz}_{i, t+1}^r & = \hat{\bz}_{i, 0}^r -  (t+1)\eta  \left( \nabla f_i(\bz_{i,0}^r)+ \bc_i^r\right) \\
&\approx P_{\teta}(\overline{\bx}^r) - (t+1)\eta  \nabla f(P_{\teta}(\overline{\bx}^r) ),    
\end{aligned}    
\end{equation}
where we substitute $ \hbz^r_{i,0}= P_{\teta}(\overline{\bx}^r)$. 
By the definition ${\bz}_{i,t+1}^r=P_{(t+1)\eta}(\hat{\bz}_{i,t+1}^r)$, we have 
\begin{equation}\label{eq-rev-8}
  {\bz}_{i,t+1}^r\approx  P_{(t+1)\eta} \Big( P_{\teta}(\overline{\bx}^r) - (t+1)\eta  \nabla f(P_{\teta}(\overline{\bx}^r) ) \Big).   
\end{equation} 

Comparing with \eqref{eq-rev-2}, we see that using $(t+1)\eta$ allows \eqref{eq-rev-8} to be close to the centralized PGD.   
}

This parameter $(t+1)\eta$  significantly improves the practical performance of Algorithm~\ref{alg-fl}, as we will demonstrate in numerical experiments.  

\section{ Analysis} \label{sec:analysis} 
In this section, we establish convergence guarantees for Algorithm \ref{alg-fl}. 
To facilitate the analysis, we impose the following assumptions on the loss function.  

\begin{assumption}\label{assm:g}
The function $g: \mathbb{R}^{d} \mapsto \mathbb{R}$ is proper closed convex, but not necessarily smooth. {
There exists a constant \( 0 < B_g < \infty \), independent of \( \bx \), such that for any \( \bx \in \mathbb{R}^d \) and any \( \widetilde{\nabla} g(\bx) \in \partial g(\bx) \), it holds that $\|\widetilde{\nabla} g(\bx)\|\le B_g$.
}
\end{assumption}
This subgradient boundedness assumption is common in analyses of non-smooth problems, and is satisfied by regularizers such as the $\ell_1$-norm, the $\ell_{\infty}$-norm, and the ReLU (Rectified Linear Unit), among others~\cite{9186389}.

\begin{assumption}\label{asm-smooth} 
Each loss function $f_i: \mathbb{R}^{d} \mapsto \mathbb{R}$ is $L$-smooth, i.e., there exists a constant $0<L<\infty$ such that 
\begin{equation*}
f_i(\by)\le f_i(\bx)+\langle \nabla f_i(\bx), \by-\bx\rangle+\frac{L}{2}\|\bx-\by\|^2, ~\forall \bx, \by\in \mathbb{R}^d.     
\end{equation*}
\end{assumption}
\begin{assumption}\label{asm-cvx}
The loss function $f$ or $F$ {defined in \eqref{eqn:basic_opt}} satisfies one of the following conditions	
\begin{enumerate}
\item[(1)]  $f$ is general non-convex.
\item[(2)]  $F$  satisfies proximal PL inequality: $	\mu\left(F(\bx)-F^{\star}\right) \leq \frac{1}{2} D_g(\bx, \mu)$,  where $\mu>0$,
$D_g(\bx, \mu):=-2 \mu \min _{\by}~$ $ \langle\nabla F(\bx), \by-\bx\rangle+\frac{\mu}{2}\|\by-\bx\|^2+g(\by)-g(\bx)$, and $F^{\star}$ is the optimal value of problem \eqref{eqn:basic_opt} \cite{j2016proximal,karimi2016linear}. 
\end{enumerate}
\end{assumption}
We will establish separate convergence guarantees for these two cases. 
%
%
The proximal PL inequality extends the PL inequality \cite{bianchin2022online,michael2023gradient} to non-smooth problems and offers a concise proof of linear convergence rates for convex and non-convex composite problems \cite{karimi2016linear}. It applies to various cases such as $\ell_1$-regularized least squares problem \cite{karimi2016linear}.

To handle the stochasticity caused by random sampling $ \mB_{i,t}^r$,  we let $\mathcal{F}_t^r$ denote the event generated 
by  $\{\mB_{i,\tilde{t}}^{\tilde{r}}~ |~ i\in [n]; \tilde{r}\in [r]; \tilde{t} \in [t]-1\}$ 
and make the following assumptions regarding the stochastic gradients.

\begin{assumption}\label{asm-sgd}
The stochastic gradients of each client $i$ satisfy
\begin{equation}
\begin{aligned}
&\bE\left[\nabla f_i(\bz_{i,t}^{r}; \mB_{i,t}^r) |  \mathcal{F}_t^r\right]=\nabla f_i(\bz_{i,t}^{r}), \\
&\bE \left[\left\|\nabla f_i(\bz_{i,t}^{r}; \mB_{i,t}^r) -\nabla f_i(\bz_{i,t}^{r})  \right\|^2 |  \mathcal{F}_t^r\right] \le \frac{\sigma^2}{b}. 
\end{aligned}
\end{equation}
\end{assumption}
Assumption \ref{asm-sgd} shows that 
the local stochastic gradients are unbiased estimates of the local full gradients and that the variance is inversely proportional to the batch size $b$.

With these assumptions in place, we can analyze the general non-convex case using the {auxiliary} function
\begin{equation}\label{eq:omega_ncvx}
\begin{aligned}
\Omega^r:= F(P_{\teta}(\ox^{r})){-F^{\star}}+ \frac{1}{n\teta }\|\bLam^r-\obLam^r\|^2,
\end{aligned}
\end{equation}
where $ \bLam^r:={\eta}( \tau\nabla \mathbf{f}({P_{\teta}(\obx^r)})+ \sum_{t=0}^{\tau-1} \overline{\nbf}(\bZ_t^{{r-1}} ; \mB_{t}^{r-1})-  \sum_{t=0}^{\tau-1} \nbf (\bZ_t^{{r-1}} ; \mB_{t}^{r-1}))$, $\obLam^r\!:=\!\col\left\{ \tfrac{1}{n} \sum_{i=1}^{n}\Lambda_i^r \right\}_{i=1}^n$ with $\Lambda_i^r  $ as the $i$-th block of $\bLam^r$ such that $\bLam^r=\col\{ \Lambda_i^r\}_{i=1}^n$, and $F^{\star}$ is the optimal value of problem \eqref{eqn:basic_opt}. The first component of $\Omega^r$ measures the suboptimality of the global model $P_{\tilde{\eta}}(\ox^r)$, while the  second component bounds the client-drift error, i.e., how far the local models $\{\bz_{i,\tau}^r\}_{i}$ are from the common initial point $P_{\teta}(\ox^r)$ after the local updates. This drift error can be bounded by the inconsistency of the local directions accumulated through the local updates, as characterized by $\|\bLam^r-\obLam^r\|^2$.

The convergence proof is based on a descent lemma for the {auxiliary} function defined in ~\eqref{eq:omega_ncvx}. Analogous to using the gradient norm as a descent magnitude in smooth non-convex problems,  our analysis measures the descent by the norm of  $\mathcal{G}(P_{\teta}(\ox^r))$ defined via
\begin{align}
\mathcal{G}(P_{\teta}(\ox^r)):=&\ \frac{1}{\teta}	
(P_{\teta}(\ox^r)-\tilde{\bx}^{r+1})\label{eq-g7}\\
{\in}&\ {\nabla f(P_{\teta}(\ox^r))+\partial g(\tilde{\bx}^{r+1}),}\label{eq-g8} 
\end{align}  
where $\tilde{\bx}^{r+1}$, defined in \eqref{eq:pgd}, is virtual and merely used for analysis. {The derivation of \eqref{eq-g8} is given in Appendix \ref{app-derive-g8}. When $g=0$, $\mathcal{G}(P_{\teta}(\ox^r))$ reduces to $\nabla f(P_{\teta}(\ox^r))$. }

In the centralized setting, $\mathcal{G}(P_{\teta}(\ox^r))$ is a commonly used metric to evaluate the first-order optimality for non-convex composite problems~\cite{j2016proximal}.
{The intuition of using the norm of \(\mathcal{G}(P_{\teta}(\ox^r))\) as a metric is that \(\mathcal{G}(P_{\teta}(\ox^r))\) can be approximately viewed as an element of the subgradient of \(f+g\), except that \(\nabla f\) and \(\partial g\) are evaluated at two different points. Moreover, $\mathcal{G}(P_{\teta}(\ox^r))={\bf 0}_d$ if and only if $ P_{\teta}(\ox^r) =\tldx^{r+1}$ and ${\bf 0}_d\in \nabla f(P_{\teta}(\ox^r)) + \partial g(P_{\teta}(\ox^r))$, which means that the global model $P_{\teta}(\ox^r)$ satisfies the first-order optimality conditions for~\eqref{eqn:basic_opt}. }
With $\Omega^r$ and $\mathcal{G}(P_{\teta}(\ox^r))$ defined, we can present the following theorem.   
\begin{theorem}[General non-convex case] \label{thm:noncvx}
Under Assumptions \ref{assm:g}, \ref{asm-smooth}, \ref{asm-cvx}-(1),   
\ref{asm-sgd}, if the step sizes satisfy 
\begin{equation}\label{eq:step-sizes}
\teta:=\eta\eta_g \tau\le \frac{1}{10L }, ~ \eta_g\ge \max\left\{ 1.5,\sqrt{\frac{n}{8}} \right\},
\end{equation}
then the sequence of $\{ \mathcal{G}(P_{\teta}(\ox^r))\}_r $ generated by the iterates of Algorithm \ref{alg-fl} satisfies
\begin{equation*}
\begin{aligned}
\frac{1}{R} \sum_{r=1}^R\bE \| \mathcal{G}(P_{\teta}(\ox^r))\|^2 \le \frac{\bE\left[\Omega^{1}\right]}{0.3\teta R}  \!+\!  \frac{20\sigma^2}{n\tau b}\!+\!\frac{187L^2\teta^2 B_g^2}{\eta_g^2}.
\end{aligned}
\end{equation*}
\end{theorem}
{Theorem 3.5 demonstrates that Algorithm 1 converges sublinearly to a residual error of order $\mathcal{O}({\sigma^2}/{(n\tau b)} + {L^2 \teta^2 B_g^2}/{\eta_g^2})$. The first term in the residual is dictated by the stochastic gradient variance $\sigma^2$ and can be mitigated by employing a large batch size $b$. If full local gradients are used, then $\sigma^2 = 0$, rendering this residual term zero. The second term in the residual stems from the bound of the subgradient $\partial g$. As this term is proportional to the square of the step size $\teta$, a smaller step size can effectively reduce this residual. 
}

When $F$ satisfies the proximal PL inequality, we can achieve the following stronger convergence guarantee:

\begin{theorem}[proximal PL inequality case]\label{thm:pl} Under Assumptions \ref{assm:g}, \ref{asm-smooth}, \ref{asm-cvx}-(2), and 
\ref{asm-sgd},  if  the step sizes satisfy \eqref{eq:step-sizes}, the sequence $\{\Omega^r\}_r$ generated by the iterates of Algorithm \ref{alg-fl}   satisfies
\begin{equation*}
\begin{aligned}
\bE [\Omega^{R+1}]\le& \left(1-\frac{\mu \teta}{3}\right)^R  \bE[\Omega^1] +  \frac{18\sigma^2}{\mu n\tau b} +   \frac{168 L^2\teta^2 B_g^2}{\mu \eta_g^2}.
\end{aligned}
\end{equation*}
\end{theorem}

Compared to the sublinear rate of Theorem \ref{thm:noncvx}, Theorem \ref{thm:pl} establishes  linear convergence to a residual error of order $\mathcal{O}({\sigma^2}/{( \mu n\tau b)}+ {  L^2\teta^2  B_g^2}/{(\mu\eta_g^2}))$. The convergence guarantee in Theorem \ref{thm:pl} is the stronger of the two, because it proves convergence to a neighborhood of the optimal loss value whereas Theorem \ref{thm:noncvx} only ensures convergence to a neighborhood of a first-order stationary point. 
{Just like in Theorem~\ref{thm:noncvx}, the first term of the residual error in Theorem 3.6 is associated with the stochastic gradient variance $\sigma^2$ and can be reduced by increasing the batch size $b$. The second term of the residual error is related to the bound of the subgradient $\partial g$ and can be decreased by using a smaller step size $\teta$.}

{ 
Neither Theorems~\ref{thm:noncvx} or \ref{thm:pl} imply the convergence of the global model $P_{\teta}(\ox^r)$ to the optimal solution set $\mathcal{X}^{\star} := \operatorname{argmin}_{\bx} F(\bx)$.}
{This is a common issue for non-convex composite optimization with first-order methods~\cite{ghadimi2016mini, j2016proximal}. Even if $\|\mathcal{G}(P_{\teta}(\ox^r))\|^2$ or $F(P_{\teta}(\ox^r)) - F^{\star}$ is small, the distance of $P_{\teta}(\ox^r)$ to $\mathcal{X}^{\star}$ can still be large. To ensure that the distance of $P_{\teta}(\ox^r)$ to $\mathcal{X}^{\star}$ is upper bounded by $\|\mathcal{G}(P_{\teta}(\ox^r))\|^2$ or $F(P_{\teta}(\ox^r)) - F^{\star}$, the objective functions typically need to be strongly convex, or satisfy other local error bound conditions~\cite{luo1993error, yue2019family}.}


\begin{remark}
In our conference publication \cite{zhang2023composite}, we investigated Algorithm \ref{alg-fl} for solving problem \eqref{eqn:basic_opt} when $f(\bx)$ is $\mu_{sc}$-strongly convex where $\mu_{sc}>0$ is a parameter. By using an  auxiliary function on the form
\begin{equation*}
\Omega_{sc}^{r}:=  \left\|P_{\teta}(\ox^{r})-\bx^{\star}\right\|^2+ \frac{1}{n}\|\bLam^r-\obLam^r\|^2, 
\end{equation*}
where $\bx^{\star}$ is the optimal solution, 
we proved linear convergence for the point sequence $\{\|P_{\teta}(\ox^{r})-\bx^{\star}\|^2\}_r$. For completeness, we include the main theorem below.
\begin{theorem}{{\bf(Strongly convex case \cite[Thm.~3.4]{zhang2023composite})}}\label{thm-sc}
Under Assumptions \ref{assm:g}, \ref{asm-smooth}, \ref{asm-sgd}, if $f(\bx)$ is $\mu_{sc}$-strongly convex and the step sizes satisfy 
\begin{equation*}
\begin{aligned}
\teta:=\eta\eta_g \tau \le \frac{\mu_{sc}}{150L^2}, ~\eta_g= \sqrt{n},
\end{aligned}
\end{equation*} 
then the sequence $\{\Omega_{sc}^r\}_r$  satisfies
\begin{equation*}
\begin{aligned}
\vspace{-2mm}
\bE\left[\Omega_{sc}^{R+1}\right]\le \left(1-\frac{\mu_{sc}\teta}{3}\right)^R\!\bE[\Omega^{1}_{sc}] +\frac{30\teta\sigma^2}{\mu_{sc} n \tau b}+\frac{21\teta B_g^2}{\mu_{sc}\eta_g^2 }.
\end{aligned}
\end{equation*}
\end{theorem}
Although we have also proven linear convergence under the proximal PL inequality without invoking convexity, the convergence of the point sequence $\{\|P_{\teta}(\ox^{r})-\bx^{\star}\|^2\}_r$ is stronger than the convergence of the loss value sequence $\{F(P_{\teta}(\ox^{r}))-F^{\star}\}_r$. 
This stronger convergence in the strongly convex case allows us to get rid of the $B_g$-term in the special case that 
$g$ is an indicator function of a convex set and $\nabla f(\bx^{\star})=0$ \cite[Corollary 3.6]{zhang2023composite}. In the non-convex cases considered in this paper, proving convergence of $\|P_{\teta}(\ox^{r})-\bx^{\star}\|$ is challenging and we have not been able to eliminate the $B_g$-term. However, in our numerical experiments, we observe that when we eliminate $\sigma^2$ by use of full gradients, our algorithm converges exactly and is unaffected by $B_g$ (Fig.~\ref{fig_1}). This suggests that there is room for improvement in our theoretical analysis. 
\end{remark}

Next, we proceed to validate the theoretical results in numerical experiments.

\section{Numerical experiments}\label{sec:experiment}
In the numerical experiments, we consider two problems: training of sparse logistic regression on synthetic heterogeneous datasets which satisfies proximal PL inequality~\cite[Section 4.1]{karimi2016linear} and training of a {convolutional} neural network classifier on the real heterogeneous dataset  MNIST~\cite{noble2022differentially} which is general non-convex. We compare our algorithm with the state-of-the-art algorithms FedMid~\cite{yuan2021federated}, FedDA~\cite{yuan2021federated}, and Fast-FedDA~\cite{bao2022fast}, which all use a fixed number of local updates to solve the composite FL problem. 

\subsection{Sparse logistic regression}
Consider the sparse logistic regression problem 
\begin{equation*}
\underset{{\bx} \in \mathbb{R}^{d}}{\operatorname{minimize}}~  \vartheta\|{\bx}\|_1+\frac{1}{n}\sum_{i=1}^{n} \frac{1}{m_i}\sum_{l=1}^{m_{i}}  \ln \left(1\!+\!\exp\left(-b_{i l}\mathbf{a}_{i l}^{T} {\bx}\right) \right),
\end{equation*}
where client $i$ owns $m_i$ training samples $(\mathbf{a}_{i l}, b_{i l}) \in \mathbb{R}^{d} \times\{-1,+1\}$ with $l=1, \ldots, m_{i}$ and $\vartheta>0$ is a regularization parameter.
We measure the performance by
$$\textit{optimality}:=\|\mathcal{G}(P_{\teta}(\ox^r))\|/ \|\mathcal{G}(P_{\teta}(\ox^1))\|, $$
which represents the relative error between $P_{\teta}(\ox^{r})$ and the first-order stationary points of  \eqref{eqn:basic_opt}. 
To generate data, we use the method in \cite{li2020federated-fedprox} which allows us to control the degree of heterogeneity by two parameters $(\alpha, \beta)$.

In the first set of experiments, we show that our algorithm can overcome client drift. 
To eliminate the influence of random sampling noise, we use the full gradients.
We set $(\alpha, \beta)=(50, 50)$, $n=30$,  $d=20$,  $\vartheta=0.003$,  $m_i=100, \forall i$. 
For the proposed algorithm, we use hand-tuned step sizes  $\eta=4$ and $\eta_g=15$. For  FedDA, we use the same step sizes $\eta=4$ and $\eta_g=15$. For FedMid, we use $\eta=1$ and $\eta_g=5$, based on the observation that smaller step sizes yield better performance. For Fast-FedDA, we use the adaptive step sizes as specified in~\cite{bao2022fast}, which are decaying step sizes. 
We set the number of local updates as $\tau\in\{1,10 \}$. 

\begin{figure}[htbp]
\begin{minipage}[h]{.9\linewidth}
\centering
\centerline{\includegraphics[width=5.5cm]{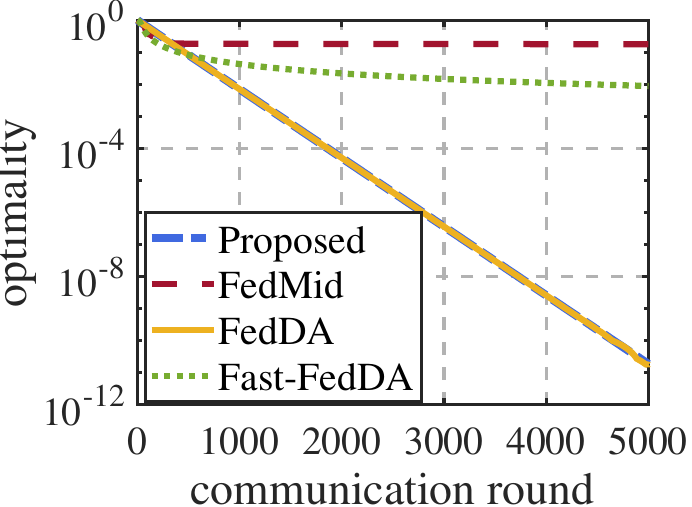}}
\end{minipage}
\hfill
\vspace{2mm}
\begin{minipage}[h]{0.9\linewidth}
\centering
\centerline{\includegraphics[width=5.5cm]{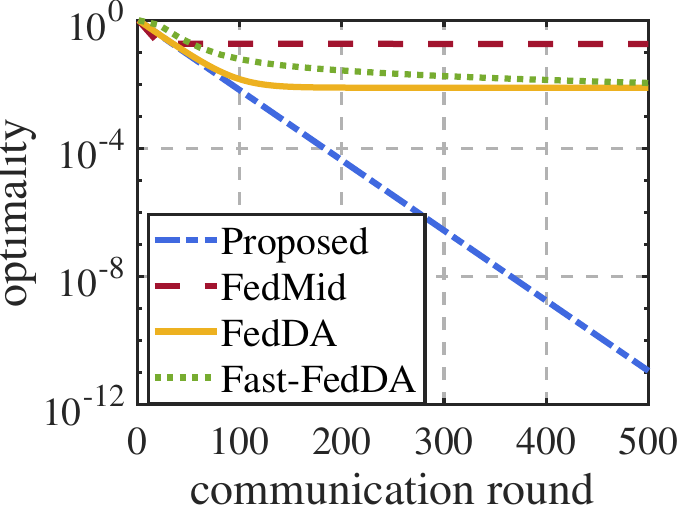}}
\end{minipage}
\caption{Sparse logistic regression using full gradients with $\tau=1$ (left) and $\tau=10$ (right), respectively. }
\label{fig_1}
\end{figure}

The results are shown in Fig.~\ref{fig_1}. 
When $\tau=1$, there are no local updates and hence  no client drift. 
Both FedDA and our algorithm converge to machine precision. The same step size setting ensures that both algorithms converge at exactly the same rate.  
Although our theoretical results suggest the existence of a residual determined by $B_g$ (the subgradient bound), our experimental results do not show any such error, indicating that there is the possibility to improve the analysis. 
FedMid performs the worst because it faces the curse of primal averaging. Fast-FedDA converges slowly due to its decaying step sizes.  
When $\tau=10$, data heterogeneity and local updates cause client drift. Our algorithm still achieves precise convergence and requires fewer communication rounds (roughly reduced by $1/\tau$) than the others. FedDA, due to client drift, only converges to a neighborhood of the first-order stationary points and can no longer maintain a similar performance as our algorithm. FedDA and FedMid perform even worse due to client drift. 

\begin{figure}[htbp]
\begin{minipage}[h]{.9\linewidth}
\centering
\centerline{\includegraphics[width=5.5cm]{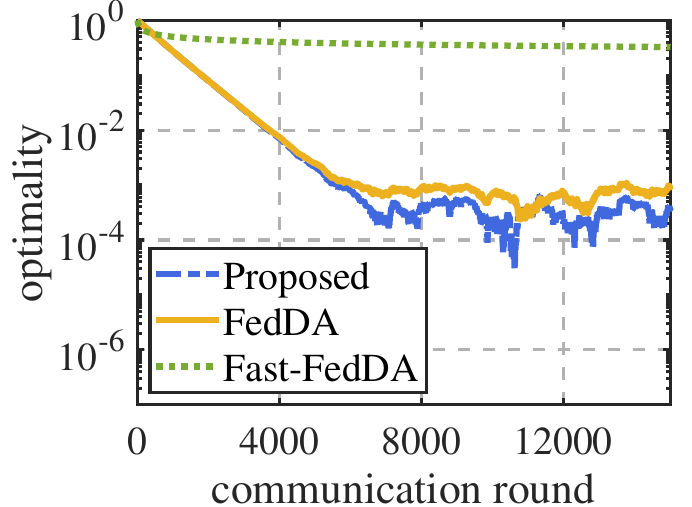}}
\end{minipage}
\hfill
\vspace{2mm}
\begin{minipage}[h]{.9\linewidth}
\centering
\centerline{\includegraphics[width=5.5cm]{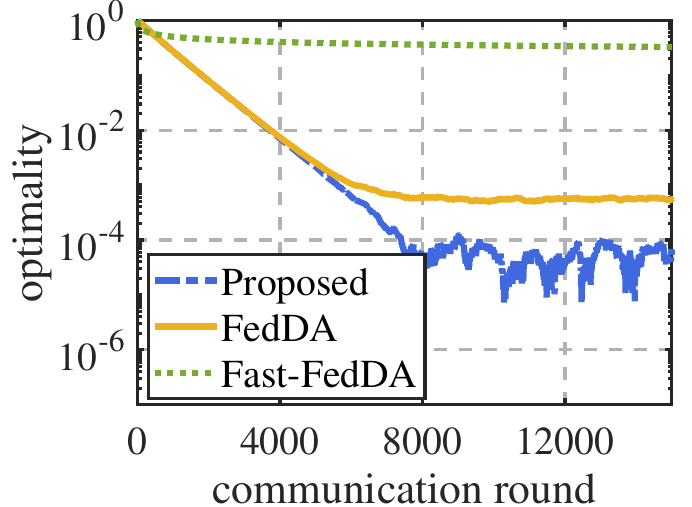}}
\end{minipage}
\caption{Sparse logistic regression using stochastic gradients with $b=1$ (left) and $b=20$ (right), respectively. }
\label{fig_2}
\end{figure}

In the second set of experiments, we use stochastic gradients and compare our algorithm with FedDA~\cite{yuan2021federated} and Fast-FedDA~\cite{bao2022fast}. We set $\vartheta=0.0005$, $m_i=2000, \forall i$,  $\tau=20$, $\eta=2$, $\eta_g=8$, $b\in \{1,20 \}$. {In the \((t+1)\)-th local update of \(r\)-th communication round, client $i$ randomly select a subset data \(\mB_{i,t}^r\) with size $b$ from 2000 data samples \( \{ \mathcal{D}_{il} := ({\bf a}_{il}, b_{il})\}_{l}\). The mini-batch stochastic gradient is then computed as \(\nabla f_i(\cdot; \mB_{i,t}^r) = \frac{1}{b} \sum_{\mathcal{D}_{il} \in \mB_{i,t}^r} \nabla f_{il}(\cdot; \mathcal{D}_{il})\).} The other settings are the same as those in Fig.~\ref{fig_1}.
As shown in Fig.~\ref{fig_2}, when we use stochastic gradients, our algorithm also converges to a neighborhood. This neighborhood is induced by the variance of stochastic gradients. When we increase the batch size from 1 to 20, the convergence precision of our algorithm improves.  
The other algorithms perform worse due to client drift and/or the use of decaying step sizes. 

\subsection{Federeated convolutional neural network training}
In the third set of experiments, we tackle the non-convex and non-smooth problem of {training a convolutional neural network (CNN) for classifying the MNIST dataset \cite{zhang2024secure}.}

{The CNN architecture consists of two convolutional layers, each with 32 feature maps and a kernel size of 3x3, followed by a 2x2 max pooling layer. The network is then connected to three fully connected layers with 64, 32, and 10 units, respectively. We use Relu activation functions in the hidden layers and a softmax activation function in the output layer. The total number of parameters is $d=112,394$.}
The training loss function is cross-entropy with a regularization term defined as $g(\bx)=\vartheta\|\bx\|_1$.

The MNIST dataset consists of grayscale images of handwritten digits, each with dimensions of $28\times28$ pixels, encompassing 10 classes that represent the digits from 0 to 9. 
{Our training dataset comprises 60,000 samples.
To introduce data heterogeneity, we uniformly sample 30,000 samples from the complete training dataset, allocating 3,000 samples to each of the 10 clients. The remaining 3,0000 samples are distributed by assigning the samples of label $l$ to client $l+1$. Note that the total number of training samples for each client may differ.  We have 10,000 test samples to evaluate the classification accuracy of the global model on the server.}

We compare the proposed algorithm with FedDA which performs best among all compared algorithms in previous experiments. { For both algorithms, we set $\vartheta=10^{-4}$, $\eta=0.005$, $\eta_g=1$, $b=10$, and $\tau\in \{5,10\}$. }

\begin{figure}[htbp]
\centering
\centerline{\includegraphics[width=5.7cm]{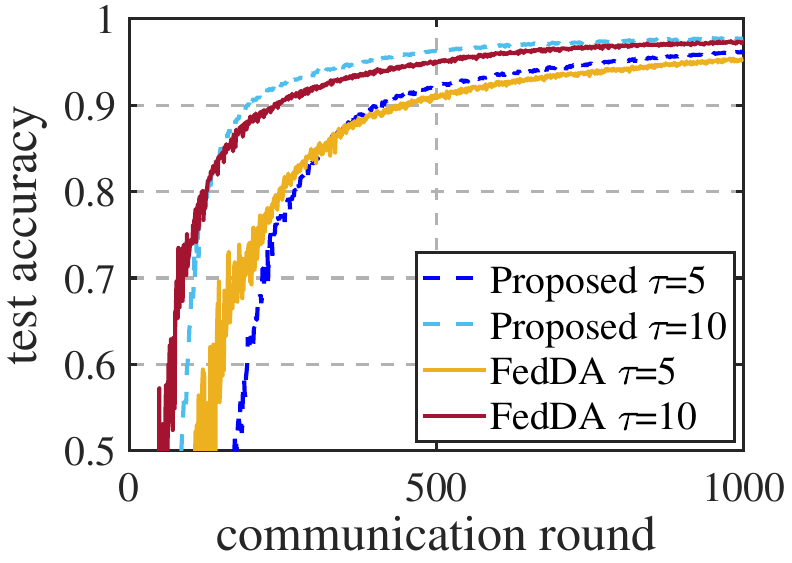}}
\caption{Classification of MNIST using CNN with $\tau=5$ and $\tau=10$.}
\label{fig_dnn}
\end{figure}

{Fig.~\ref{fig_dnn} shows the test accuracy with respect to communication rounds.  
When $\tau=5$ and $\tau=10$, our algorithm overcomes client-drift issues and achieves higher accuracy with fewer communication rounds than FedDA. }

\section{Conclusions}
We have proposed a novel algorithm tailored for federated learning with non-convex composite loss functions and heterogeneous data. 
The proposed algorithm handles non-smooth regularization by decoupling 
the proximal operator evaluation and communication. In addition, it reduces the communication frequency through local updates, exchanges only a $d$-dimensional vector per communication round per client, and effectively addresses the issue of client drift. Without imposing limitations on data similarity, we establish sublinear and linear convergence rates for the cases when the problem is generally non-convex and satisfies the proximal PL inequality, respectively. Our numerical experiments show the superiority of the proposed algorithm over the state-of-the-art on both synthetic and real datasets.


%
\appendix
\section{Appendix}\label{section-app}
\subsection{Detailed derivation of Equation~\eqref{eqn:illustration}}\label{app:prf-illustration}

In this section, we will prove that the per-client implementation of Algorithm~\ref{alg-fl} can be represented in the compact form (\ref{eqn:illustration}).
To this end, we start by writing the updates of Algorithm~\ref{alg-fl} in compact form using the notation introduced in Section~\ref{sec:intuition}
\begin{equation}\label{eqn:compact}
\hspace{-3mm}
\left\{
\begin{aligned}
&\hbZ_{t+1}^r=\hbZ_t^r-\eta \big(\nabla \mathbf{f}\left(\bZ_t^r ; \mB_{t}^r\right) +{\bf C}^r  \big), \\
&\bZ^r_{t+1}= {P_{(t+1)\eta}}\left(\hbZ^{r}_{t+1} \right),\\
&\obx^{r+1}=P_{\teta}(\obx^r)+\eta_g\left(\bW\hbZ^{r}_{\tau} - {P_{\teta}(\obx^r)}\right),\\
&{\bf C}^{r+1}=\frac{1}{\eta_g\eta\tau}( P_{\teta}(\obx^{r})\!-\!{\obx^{r+1}}) \!-\! \frac{1}{\tau}\! \sum_{t=0}^{\tau-1}\! {\nabla \mathbf{f}} \left(\bZ_t^{r} ; \mB_{t}^{r}\right),
\end{aligned}
\right.
\end{equation}
where $t\in[\tau]-1$, $r\in[R]$, and  ${\bf C}^r:=\col\{\bc_i^r\}_{i=1}^n$.

We first establish the fact that $\mathbf{W}\mathbf{C}^{r+1}=\mathbf{0}_{nd}$ for all $r$. By the initialization ${\bf C}^1=\mathbf{0}_{nd}$, this claim holds for $r=0$. Moreover, the per-client updates  \eqref{eqn:compact} imply that
\begin{equation}
\label{eq:wc}
\begin{aligned}
&\mathbf{W}\mathbf{C}^{r+1}\\ =&
\frac{1}{\eta_g\eta\tau}(P_{\tilde{\eta}}(\obx^r)-\obx^{r+1})- \frac{1}{\tau}\sum_{t=1}^{\tau-1}
\onbf(\bZ_t^r;\mB_{t}^r)\\
= &
-\frac{1}{\eta\tau}\left(\bW\widehat{\bZ}^r_{\tau}-P_{\tilde{\eta}}(\obx^r)\right)- \frac{1}{\tau}\sum_{t=1}^{\tau-1}
\onbf(\bZ_t^r;\mB_t^r),
\end{aligned}
\end{equation}
where the first equality follows from the definition of the block-wise averaging matrix $\mathbf{W}$ and that $\overline{\mathbf{X}}^r=\col\{ \overline{\bx}^r\}_{{i=1}}^n$, while the second equality uses the third step in \eqref{eqn:compact}.

By repeating the first step in \eqref{eqn:compact}, we obtain 
\begin{align}\label{eq:z_tau}
\hbZ_{\tau}^{r}& = \hbZ_{0}^{r}-\eta \Big(\sum_{t=0}^{\tau-1}{\nabla \mathbf{f}}\left(\bZ_t^{r} ; \mB_{t}^{r}\right)
+ \tau {\bf C}^r\Big).
\end{align}
Using \eqref{eq:z_tau} and 
$\hbZ_{0}^{r}=P_{\tilde{\eta}}(\obx^r)$ in 
\eqref{eq:wc} yields 
\begin{equation}\label{c=c}
\bW{\bf C}^{r+1}= \bW {\bf C}^r. 
\end{equation}
Since $\mathbf{W}\mathbf{C}^{1}=\mathbf{0}_{nd}$, it holds that
$\bW{\bf C}^r=\mathbf{0}_{nd}$ for all $r\geq 1$. 

Next, to show equivalence between the updates \eqref{eqn:compact} and \eqref{eqn:illustration}, we first note that
\begin{align*}
{\bf C}^r &= \frac{1}{\eta_g\eta\tau}
(P_{\tilde{\eta}}(\obx^{r-1})-\obx^r)
-\frac{1}{\tau}\sum_{t=0}^{\tau-1} \nbf(\bZ_t^{r-1};\mB_t^{r-1})\\
&= -\frac{1}{\eta\tau} (\bW\hat{\bZ}_{\tau}^{r-1}-P_{\tilde{\eta}}(\obx^{r-1})) - \frac{1}{\tau}\sum_{t=1}^{\tau-1} \nbf(\bZ_{t}^{r-1};\mB_t^{r-1})\\
&= \frac{1}{\tau}\!\sum_{t=1}^{\tau-1} \onbf(\bZ_{t}^{r-1};\mB_t^{r-1}) \!+\!\bW {\bf C}^{r-1}\!-\!\frac{1}{\tau}\!\sum_{t=1}^{\tau-1} \nbf(\bZ_t^{r-1};\mB_t^{r-1})\\
&= \frac{1}{\tau}\sum_{t=0}^{\tau-1} \onbf(\bZ_{t}^{r-1};\mB_t^{r-1})-\frac{1}{\tau}\sum_{t=0}^{\tau-1} \nbf(\bZ_t^{r-1};\mB_t^{r-1}).
\end{align*}
Here, we use \eqref{eq:z_tau} in the third equality and \eqref{c=c} in the last equality. 
Thus, the per-client implementation results in the first update of~\eqref{eqn:illustration}.

Next, we establish the $\obx^{r+1}$-update in 
\eqref{eqn:illustration}. The per-client implementation \eqref{eqn:compact} yields
\begin{equation*}
\begin{aligned}
\obx^{r+1} &= P_{\tilde{\eta}}(\obx^r)
+ \eta_g(\bW\widehat{\bZ}^r_{\tau} - P_{\tilde{\eta}}(\obx^r))
\\
&= P_{\tilde{\eta}}(\obx^r) -\eta\eta_g \bW\left(\sum_{t=1}^{\tau-1} \nbf(\bZ_t^r;\mB_t^{r}) +\tau  {\bf C}^r\right)\\
&= P_{\tilde{\eta}}(\obx^r) -\eta\eta_g \sum_{t=1}^{\tau-1} \onbf(\bZ_t^r;\mB_t^{r}),
\end{aligned}
\end{equation*}
where the second equality uses \eqref{eq:z_tau} and the last uses \eqref{c=c}.  We have thus verified the last step in \eqref{eqn:illustration}. 

Since the $\bZ_{t+1}^{r}$-update is the same in the two descriptions, we have shown that all updates are equivalent, and that~\eqref{eqn:illustration} captures the evolution of the per-client implementation.

\subsection{
Novel parameter selection during local updates
}\label{app:line 11}

\begin{algorithm}[t]
\caption{Proposed algorithm under the special case}
\label{alg-fl-appendix}
\begin{algorithmic}[1]
\State $\textbf{Input:}$ $R$, $\tau$, $\eta$, $\eta_g$
\State Set $\teta=\blue{\tau} \eta \eta_g$
\State Set $\ox^{1}=\bx^{\star}-\teta \nabla f_1(\bx^{\star}) $
\For {$r = 1, 2, \ldots, R$}
\State {{\bf Client} $i$}
\State Set $\hbz_{1, 0}^r=P_{\teta}(\ox^r)=\bx^{\star}$ and $\bz_{1,0}^r=P_{\teta}(\ox^r)=\bx^{\star}$
\State {${\bf t}=0$}
\State Update 
$
\begin{aligned}
&\hbz_{1,1}^r= 
\hbz_{1,0}^r-\eta \nabla f_1(\bz_{1,0}^r)=x^{\star}\!-\!\eta \nabla f_1(\bx^{\star})
\end{aligned}
$
\State Update $\bz_{1,1}^r=P_{\eta}{\left(\hbz_{1,1}^r\right)}=\bx^{\star} $    
\State {${\mathbf t}=1$}
\State Update 
$
\begin{aligned}
&\hbz_{1,2}^r= 
\hbz_{1,1}^r\!-\!\eta \nabla f_1(\bz_{1,1}^r)=\bx^{\star}\!-\!2\eta \nabla f_1(\bx^{\star})
\end{aligned}
$
\State Update $\bz_{1,2}^r=P_{2\eta}{\left(\hbz_{1,2}^r\right)}=\bx^{\star} $

{ 
\State  \hspace{0.25cm} $\vdots$
\State {${\mathbf t}=\tau-1$}
\State Update 
$
\begin{aligned}
&\hbz_{1,\tau}^r
=\bx^{\star}\!-\!\tau\eta \nabla f_1(\bx^{\star})
\end{aligned}
$
\State Update $\bz_{1,\tau}^r=P_{\tau\eta}{\left(\hbz_{1,\tau}^r\right)}=\bx^{\star} $
\State Send $ \hbz_{1,\tau}^r=\bx^{\star}-\tau\eta \nabla f_1(\bx^{\star})$ to the server}
\State {\bf Server}
\State Update $\ox^{r+1}=P_{\teta}(\ox^r)+\eta_g(  \hbz_{1,\tau}^r    -P_{\teta}(\ox^r))= \bx^{\star}+ \eta_g (\bx^{\star}-\tau\eta \nabla f_1(\bx^{\star})  -\bx^{\star}  ) =\bx^{\star}-\teta \nabla f_1(\bx^{\star})$ 
\State Broadcast $\ox^{r+1}=\bx^{\star}-\teta \nabla f_1(\bx^{\star})$  to the workers	
\State {{\bf Client} $i$}
\State Receive $\ox^{r+1}=\bx^{\star}-\teta \nabla f_1(\bx^{\star})$ from the server
\EndFor
\State \textbf{Output:} $P_{\teta}(\ox^{R+1})=\bx^{\star}$
\end{algorithmic}
\end{algorithm}

In Algorithm \ref{alg-fl}, Line 10 incorporates the parameter $(t+1)\eta$ in the computation of $P_{(t+1)\eta}{\left(\hbz_{i,t+1}^r\right)}$. 
This parameter is carefully chosen based on testing our algorithm under a special case:  $\operatorname{minimize}_{\bx\in \mathbb{R}^d}~f_1(\bx)+g(\bx)$, where we set $n=1$, and use full gradient.  
Notably, in this case, the correction term is always equal to ${\bf 0}_d$ due to $n=1$. It is known that any first-order stationary point $\bx^{\star}$ satisfies the condition $\bx^{\star}=P_{\beta}(\bx^{\star}-\beta \nabla f_1(\bx^{\star}))$ for any positive $\beta$. 
A well-designed algorithm should be able to ``stop'' at $\bx^{\star}$.  
The application of $(t+1)\eta$ achieves this property as elaborated in Algorithm \ref{alg-fl-appendix}.
More precisely, when the local update begins from $\bx^{\star}$, the initial value of the pre-proximal global model $\overline{\bx}^1$ is 
$\bx^{\star}-\teta \nabla f_1(\bx^{\star})$, and the output is always $\bx^{\star}$. 

{In the special case where \( n = 1 \) and we use full gradients, we have shown that employing the parameter \((t+1)\eta\) allows Algorithm 2 to stop at \( \bx^{\star} \). However, this parameter choice is heuristic, as Algorithm 1 is designed for the more general scenario where \( n \geq 2 \) and stochastic gradients are used. Our theorems indicate that Algorithm 1 is subject to residual errors due to stochastic gradients and non-smoothness, preventing it from converging exactly to \( \bx^{\star} \) and stopping at \( \bx^{\star} \).

Despite the heuristic basis for the \((t+1)\eta\) parameter, our empirical results demonstrate that this choice enhances the algorithm's performance beyond what our theoretical predictions suggest; cf. Fig. 2 (left). Moreover, our theoretical analysis is established under the use of \((t+1)\eta\) and shows that the resulting residual error is small. Therefore, while the design of \((t+1)\eta\) was initially heuristic, both our empirical and theoretical findings validate that this choice is effective and well-justified.}

\subsection{Derivation of \eqref{eq-g8}}\label{app-derive-g8}
{
According to \eqref{eq:pgd}, we have 
\begin{equation}
\tldx^{r+1}=\operatorname{argmin}_{\bx}~ \teta g(\bx) + \frac{1}{2}\| \bx- \boldsymbol{\omega}^r \|^2,   
\end{equation}
where we define $\boldsymbol{\omega}^r:=  P_{\teta}(\ox^r)-\teta \nabla f(P_{\teta}(\ox^r))$ for simplicity. 
By the optimality condition of $\tldx^{r+1}$, we have 
\begin{equation}
\frac{1}{\teta} (\boldsymbol{\omega}^r-\tldx^{r+1}) \in \partial g(\tldx^{r+1}).      
\end{equation}
Substituting $\boldsymbol{\omega}^r $ and reorganizing the result we get 
\begin{equation}
\frac{1}{\teta} (P_{\teta}(\ox^r) -\tldx^{r+1}) \in \nabla f(P_{\teta}(\ox^r)) + \partial g(\tldx^{r+1}),  
\end{equation}
which is exactly \eqref{eq-g8}. 
}

\subsection{Two preliminary lemmas}
As a first step towards proving Theorems~\ref{thm:noncvx} and~\ref{thm:pl}, we establish a bound on the drift error between the local models $\{\bz_{i,t}^r\}$ and their common initial point $P_{\teta}(\ox^r)$.
\begin{lemma}\label{lem:phi-bar-x}
Under Assumptions \ref{assm:g}, \ref{asm-smooth}, and \ref{asm-sgd}, if $\teta \le {\eta_g}/(\sqrt{20} L)$, we have
\begin{equation}\label{eq:phi--29}
\begin{aligned}
&\mathbb{E}\Big[\sum_{t=0}^{\tau-1} \sum_{i=1}^n\left\|\bz_{i,t}^r- {P_{\teta}(\ox^r)}\right\|^2\Big]\\
\le & \;
5\tau^3\eta^2 n\cdot 4 B_g^2  + 5 n \tau^3\eta^2 \|\mathcal{G}_{\teta g}( P_{\teta}(\ox^r))  \|^2 \\
& + 5 \tau  \mathbb{E}\| \bLam^r -  \obLam^r \|^2+ 10n \tau^2 \eta^2 \frac{\sigma^2}{b}. 
\end{aligned}
\end{equation}
\end{lemma}
\begin{pf}	
If $\tau=1$, then $\bz_{i,t}^r-P_{\teta}(\ox^r) ={\bf 0}_{d} $. 
For $\tau \geq 2$, repeated application of Line 9 in Algorithm~\ref{alg-fl} yields  
\begin{equation*}
\begin{aligned}
\hbz_{i,t+1}^r=\hbz_{i,0}^r-\!\eta \! \sum_{\ell=0}^{t}\!\big(\nabla f_i(\bz_{i,\ell}^r;\mB_{i,\ell}^r)+\bc_i^r \big).        
\end{aligned}
\end{equation*}
Since $\hbz_{i,0}^r=P_{\teta}(\ox^r)$ and $\bz_{i,t+1}^r = 
P_{(t+1)\eta}(\widehat{\bz}_{i,t+1}^r)$, we have
\begin{align}\label{eq-phi-P1}
&\mathbb{E} \left\|\bz_{i, t+1}^r- {P_{\teta}(\ox^r)}\right\|^2\\ \nonumber
=& \mathbb{E}\big\|P_{(t\!+\!1)\eta} \big( {P_{\teta}(\ox^r)}\!-\!\eta \! \sum_{\ell=0}^{t}\!\big(\nabla f_i(\bz_{i,\ell}^r;\mB_{i,\ell}^r)\!+\!\bc_i^r \big) \big) \!-\! {P_{\teta}(\ox^r)} \big\|^2.
\end{align}
The right-hand side of~\eqref{eq-phi-P1} can be interpreted as an ``inexact'' PGD, executed from the starting point $P_{\teta}(\ox^r)$ using the inexact gradient
$-\!\eta \! \sum_{\ell=0}^{t}\!\big(\nabla f_i(\bz_{i,\ell}^r;\mB_{i,\ell}^r)\!+\!\bc_i^r \big)$. This inspires us to bound \eqref{eq-phi-P1} using the deviation from the exact PGD step with the same initial point
$$\tilde{\bx}_{\rm pgd}:=P_{(t+1)\eta}\left({P_{\teta}(\ox^r)}-(t+1)\eta \nabla f({P_{\teta}(\ox^r)}) \right).$$ 
In this way,
\begin{align}\label{eq:2term}
&\mathbb{E} \left\|\bz_{i, t+1}^r- {P_{\teta}(\ox^r)}\right\|^2\\ \nonumber
=&  \;\mathbb{E}\big\|P_{(t\!+\!1)\eta} \big( {P_{\teta}(\ox^r)}\!-\!\eta \! \sum_{\ell=0}^{t}\!\big(\nabla f_i(\bz_{i,\ell}^r;\mB_{i,\ell}^r)\!+\!\bc_i^r \big) \big)- \tilde{\bx}_{\rm pgd}\\\nonumber
&+ \tilde{\bx}_{\rm pgd}- P_{\teta}(\ox^r)\big\|^2\\ \nonumber
\le&\;\underbrace{2\mathbb{E}\big\|P_{(t+1)\eta} \big({P_{\teta}(\ox^r)}  \!-\!\eta  \sum_{\ell=0}^{t}(\nabla f_i(\bz_{i,\ell}^r;\mB_{i,\ell}^r)\!+\!\bc_i^r ) \big)\!-\!\tilde{\bx}_{\rm pgd}\big\|^2}_{(\rm I)} \\\nonumber
&+ \underbrace{2 \bE \big\| \tilde{\bx}_{\rm pgd}  - {P_{\teta}(\ox^r)} \big\|^2}_{(\rm II)},
\end{align}
where the last step  follows from $\|a+b\|^2\le 2\|a\|^2+2\|b\|^2$. 

To bound the term (II) 
we will use the optimality condition of $\tilde{\bx}_{\rm pgd}$. By definition of the proximal operator,
\begin{align*}
\tilde{\bx}_{\rm pgd}=\operatorname{argmin}_{\bu} ~ &\frac{1}{2}\|{P_{\teta}(\ox^r)}-(t+1)\eta \nabla f({P_{\teta}(\ox^r)})-{\bf u}\|^2\\
&+(t+1)\eta g({\bf u}), 
\end{align*}
so $\tilde{\bx}_{\rm pgd}$ satisfies the optimality condition 
\begin{align*}
&(t+1)\eta (\tilde{\nabla} g(\tilde{\bx}_{\rm pgd})+ \nabla f({P_{\teta}(\ox^r)})) + \tilde{\bx}_{\rm pgd} -{P_{\teta}(\ox^r)}={\bf 0}_d,
\end{align*}
for some $\tilde{\nabla} g(\tilde{\bx}_{\rm pgd})\in \partial g(\tilde{\bx}_{\rm pgd})$. 
This implies that 
\begin{equation} \label{eq:g1}
\tilde{\bx}_{\rm pgd} - {P_{\teta}(\ox^r)} 
= -(t+1)\eta \underbrace{\left( \nabla f( {P_{\teta}(\ox^r)}) + \tilde{\nabla} g(\tilde{\bx}_{\rm pgd}) \right)}_{:=\mathcal{G}_{(t+1)\eta}( {P_{\teta}(\ox^r)})}.
\end{equation}
Note that the server step-size is $\teta$ and not $(t+1)\eta$. To correct for this, we use the iterate $\tilde{\bx}^{r+1}$ defined in \eqref{eq:pgd}, which is the same as $\tilde{\bx}_{\rm pgd}$ except that the step size $(t+1)\eta$  is replaced by $\teta$. Following similar steps as above, the optimality condition for $\tilde{\bx}^{r+1}$ implies that
\begin{equation} \label{eq:g2}  
\hspace{-2mm}
\tilde{\bx}^{r+1} - {P_{\teta}(\ox^r)} 
= -\teta \underbrace{\left( \nabla f( {P_{\teta}(\ox^r)}) + \tilde{\nabla} g(\tilde{\bx}^{r+1}) \right)}_{:=\mathcal{G}_{\teta}( {P_{\teta}(\ox^r)})}, \end{equation}
for some $\tilde{\nabla} g(\tilde{\bx}^{r+1}) \in \partial g(\tilde{\bx}^{r+1})$. We can now use~\eqref{eq:g1} and~\eqref{eq:g2} to bound (II) through the following steps
%
%
\begin{align} \label{eq-phi-P1-2}
({\rm II})=&2 \bE \|{(t+1)\eta} \mathcal{G}_{(t+1)\eta }( P_{\teta}(\ox^r) )\|^2\\ \nonumber
=& 2 \bE \big\|{(t+1)\eta} \left(\mathcal{G}_{(t+1)\eta } (P_{\teta}(\ox^r) ) -  \mathcal{G}_{\teta}(P_{\teta}(\ox^r) ) \right) \\\nonumber
&+ (t+1)\eta   \mathcal{G}_{\teta}(P_{\teta}(\ox^r) )\big\|^2\\\nonumber
=& 2 \bE \big\|{(t+1)\eta} \big(\tilde{\nabla} g(\tilde{\bx}_{\rm pgd} ) -  \tilde{\nabla} g(\tilde{\bx}^{r+1} )\big) \\\nonumber
&+ (t+1)\eta   \mathcal{G}_{\teta} (P_{\teta}(\ox^r) )\big\|^2\\\nonumber
\le& 4\tau^2\eta^2 \cdot 4 B_g^2 + 4 \tau^2\eta^2 \|\mathcal{G}_{\teta}( P_{\teta}(\ox^r))  \|^2,
\end{align}
where the last step uses Assumption~\ref{assm:g} and $t\leq \tau-1$.
%

Let us now turn our attention to bounding the term (I) on the right hand of \eqref{eq-phi-P1}. By the definition of $\tilde{\bx}_{\rm pgd}$  and the nonexpansiveness of the proximal operator, we have
\begin{align}
({\rm I})= & 2\mathbb{E}\big\|P_{(t+1)\eta} \big( {P_{\teta}(\ox^r)}  -\eta  \sum_{\ell=0}^{t}\big(\nabla f_i(\bz_{i,\ell}^r;\mB_{i,\ell}^r)+\bc_i^r \big) \big) \notag \\ 
-& P_{(t+1)\eta}\big({P_{\teta}(\ox^r)}-(t+1)\eta \nabla f({P_{\teta}(\ox^r)}) \big)\big\|^2 \label{w-22}\\
\le &  2\mathbb{E}\Big\| \eta  \sum_{\ell=0}^{t}\Big(\nabla f_i(\bz_{i,\ell}^r;\mB_{i,\ell}^r)+\bc_i^r -\nabla f({P_{\teta}(\ox^r)}) \Big)  \Big\|^2. \notag
\end{align}
The right-hand side of~\eqref{w-22} measures the difference between 
the direction $\nabla f_i(\bz_{i,\ell}^r;\mB_{i,\ell}^r)+\bc_i^r$ used by our algorithm and the exact gradient $\nabla f({P_{\teta}(\ox^r)})$. To expose how this  difference is affected by both the drift error and sampling, we will re-write it in terms of $\| \Lambda_i^r -  \overline{\Lambda}^r \|^2$ where $\overline{\Lambda}^r=\tfrac{1}{n} \sum_{i=1}^{n}\Lambda_i^r$, used in our {auxiliary} function, and the sampling variance $\sigma^2$ given in Assumption \ref{asm-sgd}. To this end, we substitute the definition of the correction term $\bc_i^r$ and obtain
\begin{align}\label{eq:I}
({\rm I})\le& \;2\mathbb{E}\Big\| \eta  \sum_{\ell=0}^{t}\Big( \nabla f_i(\bz_{i,\ell}^r;\mB_{i,\ell}^r)- \nabla f_i({P_{\teta}(\ox^r)})\\\nonumber
&+ \nabla f_i({P_{\teta}(\ox^r)}) +\frac{1}{\tau} \sum_{t=0}^{\tau-1} \frac{1}{n} \sum_{i=1}^{n} \nabla f_i (\bz_{i,t}^{{r-1}} ; \mB_{i,t}^{r-1})
\\\nonumber
&-\frac{1}{\tau} \sum_{t=0}^{\tau-1} \nabla f_i \left(\bz_{i,t}^{{r-1}} ; \mB_{i,t}^{r-1}\right) -\nabla f({P_{\teta}(\ox^r)}) \Big)  \Big\|^2\\\nonumber
=\;&2\mathbb{E}\Big\| \eta  \sum_{\ell=0}^{t}\Big( \nabla f_i(\bz_{i,\ell}^r;\mB_{i,\ell}^r)- \nabla f_i({P_{\teta}(\ox^r)})\\\nonumber
&+ \frac{1}{\eta \tau} \big(\Lambda_i^r -\overline{\Lambda}^r\big) \Big)\Big\|^2\\ \nonumber
\le &\; \underbrace{4\mathbb{E}\big\| \eta  \sum_{\ell=0}^{t}\big( \nabla f_i(\bz_{i,\ell}^r;\mB_{i,\ell}^r)- \nabla f_i({P_{\teta}(\ox^r)}) \big)\big\|^2 }_{(\rm III)} \\\nonumber
&+  4\mathbb{E}\Big\| \frac{t+1}{ \tau} \Lambda_i^r - \frac{t+1}{ \tau} \overline{\Lambda}^r \Big\|^2,
\end{align}
where we substitute $\bc_i^1={\bf 0}_d$ and $\nabla f_i(\bz_{i,t}^0;\mB_{i,t}^0)={\bf 0}_d$. 
Next, we use Assumption \ref{asm-sgd} to bound the term (III). By adding and subtracting $\nabla f_i(\bz^r_{i,\ell})$, we have
\begin{align}\label{eq:iii}
(\rm III)= & 4\mathbb{E}\big\| \eta  \sum_{\ell=0}^{t}\Big( \nabla f_i(\bz_{i,\ell}^r;\mB_{i,\ell}^r)-\nabla f_i(\bz^r_{i,\ell})\\\nonumber
&+\nabla f_i(\bz^r_{i,\ell})- \nabla f_i({P_{\teta}(\ox^r)}) \Big)\big\|^2  \\\nonumber
\le& 8 (t\!+\!1)^2\mathbb{E}\Big\|  \frac{\eta}{t+1}  \sum_{\ell=0}^{t}\Big( \!\nabla f_i(\bz_{i,\ell}^r;\mB_{i,\ell}^r)\!-\!\nabla f_i(\bz^r_{i,\ell})\Big)\Big\|^2\\\nonumber
&+8\mathbb{E}\big\|\eta  \sum_{\ell=0}^{t}\Big(\nabla f_i(\bz^r_{i,\ell})- \nabla f_i({P_{\teta}(\ox^r)}) \Big)\big\|^2.
\end{align}
The second term on the right hand of \eqref{eq:iii} can be bounded using Assumption \ref{asm-smooth}, while the first term can be handled by the fact \cite[Corollary C.1]{noble2022differentially} that
\begin{align}\label{eq:corollary} 
&{\bE}\Big\| \frac{1}{t+1} \sum_{\ell=0}^{t}   \left(\nabla f_i(\bz_{i, \ell}^r ; \mB_{i,\ell}^r)-\nabla f_i(\bz_{i, \ell}^r)\right)\Big\|^2 \\ \nonumber
= & \frac{1}{(t+1)^2} \sum_{\ell=0}^{t}  {\bE}  \Big[ {\bE} \big[ \|   \left( \nabla f_i\left(\bz_{i, \ell}^r ; \mB_{i,\ell}^r\right) \!-\! \nabla f_i\left(\bz_{i, \ell}^r\right)\right) \|^2 
| \mathcal{F}_t^r \!\big]\Big] \\ \nonumber
\le& \frac{1}{t+1} \frac{\sigma^2}{ b}. 
\end{align}
Substituting \eqref{eq:iii} and  \eqref{eq:corollary} into \eqref{eq:I}, we thus have
\begin{align}\label{eq:final I}
(\rm I)\le& 8 (t+1)\eta^2 L^2   \sum_{\ell=0}^{t}\mathbb{E}\| \bz_{i,\ell}^r-{P_{\teta}(\ox^r)} \|^2  \\
&+  4 \left(\frac{t+1}{\tau}\right)^2 \mathbb{E}\| \Lambda_i^r -  \overline{\Lambda}^r \|^2+ 8 (t+1) \eta^2 \frac{\sigma^2}{b}, \nonumber
\end{align}
where we have used the $L$-smoothness Assumption \ref{asm-smooth}. 
Now, plugging \eqref{eq:final I} and \eqref{eq-phi-P1-2} into \eqref{eq:2term} yields 
\begin{align}\label{w-23}
&\mathbb{E}[ \left\|\bz_{i, t+1}^r- {P_{\teta}(\ox^r)}\right\|^2]\\\nonumber
&\le  8 (t+1)\eta^2 L^2   \sum_{\ell=0}^{t}\mathbb{E}\| \bz_{i,\ell}^r-{P_{\teta}(\ox^r)} \|^2
+ 16\tau^2\eta^2  B_g^2 \\\nonumber
&+ 4 \tau^2\eta^2 \|\mathcal{G}_{\teta g}( P_{\teta}(\ox^r))  \|^2 + 4  \mathbb{E}\| \Lambda_i^r -  \overline{\Lambda}^r \|^2+ 8 \tau \eta^2 \frac{\sigma^2}{b}. 
\end{align}
To get a recursion for the drift error from \eqref{w-23}, we let $A^r$ denote the sum of the last four terms on the right hand of \eqref{w-23} and define  $S^r_{i,t}:=\sum_{\ell=0}^t \bE \|\bz_{i, \ell}^r- {P_{\teta}(\ox^r)}\|^2 $. In this way, $\mathbb{E}[ \left\|\bz_{i, t+1}^r- {P_{\teta}(\ox^r)}\right\|^2]= S^r_{i,t+1} -S^r_{i,t}$ and  \eqref{w-23} implies that
\begin{equation}\label{w-24}
\begin{aligned}
S^r_{i,t+1}
\le  \left(1+1/{(8\tau)}\right) S^r_{i,t} + A^r, 
\end{aligned}
\end{equation}
since $ 8 (t+1)\eta^2 L^2\le 1/(8\tau) $ when $ \teta \le {\eta_g}/(8 L) $. By repeated application of \eqref{w-24}, we conclude that 
\begin{equation}\label{drift-err-i}
\begin{aligned}
S^r_{i,\tau-1}
\leq &A^r \sum_{\ell=0}^{\tau-2}\left(1+1/{(8\tau)}\right)^{\ell} 
\le 1.15 \tau A^r,
\end{aligned}
\end{equation}
since $\sum_{\ell=0}^{\tau-2}\left(1+1/{(8\tau)}\right)^\ell \leq\sum_{\ell=0}^{\tau-2} \exp \left({ \ell}/{(8\tau)}\right) \leq\sum_{\ell=0}^{\tau-2} \exp (1/8) \le 1.15\tau$. 
Summing \eqref{drift-err-i} over all the clients $i$ completes the proof of Lemma \ref{lem:phi-bar-x}.
\end{pf}

With the  drift error bound in Lemma \ref{lem:phi-bar-x} at hand, we are ready to establish a recursion for the second part of our {auxiliary} function,
$\frac{1}{n}\bE\|\bLam^{r+1}-\obLam^{r+1}\|^2$.
\begin{lemma} \label{lem-Lambda}
Under Assumptions \ref{assm:g}, \ref{asm-smooth}, and \ref{asm-sgd}, 
if $\teta \le \frac{\eta_g}{\sqrt{20} L}$, we have	
\begin{equation*}
\begin{aligned}
&\frac{1}{n}\bE\|\bLam^{r+1}-\obLam^{r+1}\|^2-2\eta^2\tau^2 L^2 \bE\left\|P_{\teta}(\ox^{r+1})\!-\!P_{\teta}(\ox^r)\right\|^2\\
\le&\frac{1}{n} {4 \eta^2 \tau L^2} \Big(
5\tau^3\eta^2 n\cdot 4 B_g^2  + 5 n \tau^3\eta^2 \|\mathcal{G}_{\teta g}( P_{\teta}(\ox^r))  \|^2 \\
&+ 5 \tau  \mathbb{E}\| \bLam^r -  \obLam^r \|^2+ 10n \tau^2 \eta^2 \frac{\sigma^2}{b} \Big)+ \frac{1}{n} {4} \eta^2  n^2\tau^2  \frac{\sigma^2}{n\tau b}.  
\end{aligned}
\end{equation*}
\end{lemma}
\begin{pf} 	
By the definition of $\bLam^{r+1}$ and $\obLam^{r+1}$  we have 
\begin{align}\label{eq:lambda-20}
&\bE\|\bLam^{r+1}-\obLam^{r+1} \|^2\\\nonumber
=&\eta^2\bE\big\| \tau\nabla \mathbf{f}(P_{\teta}(\obx^{r+1}))-  \sum_{t=0}^{\tau-1} \nbf \left(\bZ_t^{{r}} ; \mB_{t}^r\right)\\\nonumber
&-\tau\overline{\nabla\mathbf{f}}(P_{\teta}(\obx^{r+1}))+ \sum_{t=0}^{\tau-1} \overline{\nbf}\left(\bZ_t^{{r}} ; \mB_{t}^r\right)\big\|^2 \\\nonumber
\le& \eta^2\bE\big\| \tau\nabla \mathbf{f}(P_{\teta}(\obx^{r+1}))-  \sum_{t=0}^{\tau-1} \nbf \left(\bZ_t^{{r}} ; \mB_{t}^r\right) \big\|^2,
\end{align}
where the inequality follows from the fact that
$\left\| \mathbf{Y} -  \bW\mathbf{Y}\right\|^2= \left\| \mathbf{Y} - ( \frac{1}{n}\mathbf{1}_n \mathbf{1}_n^T\otimes \mathbf{I}_d ) \mathbf{Y}\right\|^2  \le \| \mathbf{Y}\|^2$
holds for all $\mathbf{Y}$.
Next, adding and subtracting  
$\nabla \mathbf{f}(P_{\teta}(\obx^{r}))$ and $\sum_{t=0}^{\tau-1} \nabla \mathbf{f}({\bZ}_t^{r})$ to the argument of the norm  
yields 
\begin{align}\label{eq:lambda-21}
&\bE\|\bLam^{r+1}-\obLam^{r+1} \|^2\\\nonumber
\le & \eta^2 \bE \Big\|  \tau\nabla \mathbf{f}(P_{\teta}(\obx^{r+1}))- \tau\nabla \mathbf{f}(P_{\teta}(\obx^{r}))+ \tau\nabla \mathbf{f}(P_{\teta}(\obx^{r})) \\\nonumber
&- \sum_{t=0}^{\tau-1}\nabla \mathbf{f}({\bZ}_t^{r}) + \sum_{t=0}^{\tau-1}\nabla \mathbf{f}(\bZ_t^{r}) -\sum_{t=0}^{\tau-1} \nbf (\bZ_t^{{r}} ; \mB_{t}^r)\Big\|^2\\\nonumber
{\leq}&  2\eta^2\tau^2 L^2 n  {\bE\left\|P_{\teta}(\ox^{r+1})-P_{\teta}(\ox^r)\right\|^2}\\\nonumber
&+ {4}\eta^2 \tau L^2 {\sum_{t=0}^{\tau-1} \sum_{i=1}^n\bE\left\|\bz_{i,t}^r-P_{\teta}(\ox^r)\right\|^2}+ {4}\eta^2 \tau n\frac{\sigma^2}{b}. 
\end{align}
Here, the inequality follows from applying
$\|a+b\|^2\le 2\|a\|^2+2\|b\|^2 $ twice, utilizing the $L$-smoothness Assumption \ref{asm-smooth}, and employing similar derivations as in \eqref{eq:corollary} for $t+1=\tau$. 
Reorganizing \eqref{eq:lambda-21}, we get
\begin{align}\label{eq: A.13}
&\bE \|\bLam^{r+1}-\obLam^{r+1}\|^2-2\eta^2\tau^2 L^2 n  {\bE\left\|P_{\teta}(\ox^{r+1})-P_{\teta}(\ox^r)\right\|^2}\nonumber\\
\le &{4}\eta^2 \tau L^2 {\sum_{t=0}^{\tau-1} \sum_{i=1}^n\bE\left\|\bz_{i,t}^r\!-\!P_{\teta}(\ox^r)\right\|^2}\!+\! {4}\eta^2 \tau n\frac{\sigma^2}{b}.
\end{align}
By substituting the drift error given by \eqref{eq:phi--29}  into  \eqref{eq: A.13} and then multiplying both sides of the result by ${1}/{n}$, we arrive at the desired result, and the proof is complete. 
\end{pf}

\subsection{Proof of Theorem \ref{thm:noncvx}}\label{prf:thm-ncvx}
In the proofs of Lemmas \ref{lem:phi-bar-x} and \ref{lem-Lambda}, our focus was on the local update. Starting from this section, we shift our attention to the server-side update.
We begin with the following useful fact \cite[Lemma 1]{j2016proximal}:
\begin{fact}\label{fct:3point}
Let ${\bx^{+}}:=P_{\eta}(\bx-\eta \bv)$ where $\eta>0$, then we have
$$
\begin{aligned}
& F\left({\bx^{+}}\right) \leq F({\bz})+\left\langle\nabla F(\bx)-\bv, {\bx^{+}}\!-\!{\bz}\right\rangle\\
&-\frac{1}{\eta}\left\langle {\bx^{+}}-\bx, {\bx^{+}}-{\bz}\right\rangle
+\frac{L}{2}\left\|{\bx^{+}}-\bx\right\|^2+\frac{L}{2}\|{\bz}-\bx\|^2,  \forall \bz.    
\end{aligned}
$$
\end{fact}
{Here, \(\bx^+\) refers to the variable after one update starting from \(\bx\).}
Fact \ref{fct:3point} gives the function-value decrease after an update on the form 
${\bx^{+}}:=P_{\eta}(\bx-\eta \bv)$. Our analysis relies on two updates that follow such a pattern: the server-side update~\eqref{eq:bar-27} and the 
virtual centralized PGD update \eqref{eq:pgd}. We will apply Fact \ref{fct:3point} to \eqref{eq:pgd} and \eqref{eq:bar-27}, respectively.

First, applying Fact~\ref{fct:3point} to \eqref{eq:pgd} using $\bx^+=\tilde{\bx}^{r+1}$, $\bz=P_{\teta}(\ox^{r})$, $\bx=P_{\teta}(\ox^{r})$, and $\bv=\nabla f(P_{\teta}(\ox^{r}))$ yields
\begin{align}\label{eq:y22}
&\mathbb{E}\left[F\left(\tilde{\bx}^{r+1}\right)\right] \leq \mathbb{E}\Big[F\left(P_{\teta}(\ox^r)\right)\\\nonumber
&+\!\left(\frac{L}{2}-\frac{1}{2 \teta}\right)\left\|\tilde{\bx}^{r+1}-P_{\teta}(\ox^r)\right\|^2-\frac{1}{2 \teta}\left\|\tilde{\bx}^{r+1}\!-\!P_{\teta}(\ox^r)\right\|^2\!\Big].
\end{align}
Next, applying Fact \ref{fct:3point} to  \eqref{eq:bar-27} with $\bx^+=P_{\teta}(\ox^{r+1})$, $\bx=P_{\teta}(\ox^{r})$, $\bz=\tilde{\bx}^{r+1}$, and $\bv=\bv^{r}$ gives
\begin{align}\label{eq:y23}
& \mathbb{E}\left[F\left({P_{\teta}(\ox^{r+1})}\right)\right] \\\nonumber
\leq& \bE \Big[ F(\tilde{\bx}^{r+1})+\left\langle\nabla f(P_{\teta}(\ox^r))-\bv^{r}, {P_{\teta}(\ox^{r+1})}-{\tilde{\bx}^{r+1}}\right\rangle\\\nonumber
&-\frac{1}{\teta}\left\langle {P_{\teta}(\ox^{r+1})}-P_{\teta}(\ox^r), {P_{\teta}(\ox^{r+1})}-{\tilde{\bx}^{r+1}}\right\rangle\\\nonumber
&+\frac{L}{2}\left\|{P_{\teta}(\ox^{r+1})}-P_{\teta}(\ox^r)\right\|^2+\frac{L}{2}\|{\tilde{\bx}^{r+1}}-P_{\teta}(\ox^r)\|^2 \Big] \\\nonumber
=&\mathbb{E}\Big[F\left(\tilde{\bx}^{r+1}\right)+\left\langle\nabla f(P_{\teta}(\ox^r))-\bv^{r}, {P_{\teta}(\ox^{r+1})}-{\tilde{\bx}^{r+1}}\right\rangle \\\nonumber
&+\left(\frac{L}{2}-\frac{1}{2 \teta}\right)\left\|{P_{\teta}(\ox^{r+1})}-P_{\teta}(\ox^r)\right\|^2\\\nonumber
+&\!\left(\frac{L}{2}\!+\!\frac{1}{2 \teta}\right)\left\|{\tilde{\bx}^{r+1}}\!-\!P_{\teta}(\ox^r)\right\|^2\!-\!\frac{1}{2 \teta}\left\|{P_{\teta}(\ox^{r+1})}-{\tilde{\bx}^{r+1}}\right\|^2\!\Big],
\end{align}
where the last step uses  $2\langle b, b-a \rangle=\|b-a\|^2+\|b\|^2-\|a\|^2 $. We now use~\eqref{eq:y22} to eliminate the virtual function value $F(\tilde{\bx}^{r+1})$ in \eqref{eq:y23} and simplify:
\begin{align}\label{eq:f-f-25}
&\mathbb{E}\left[F\left({P_{\teta}(\ox^{r+1})}\right)\right] \\ \nonumber
\le&\mathbb{E}\big[F\left(P_{\teta}(\ox^r)\right)+\left(L-\frac{1}{2 \teta}\right)\|\tilde{\bx}^{r+1}-P_{\teta}(\ox^r)\|^2\\ \nonumber
+&\!\left(\frac{L}{2}\!-\!\frac{1}{2 \teta}\right)\|{P_{\teta}(\ox^{r+1})}\!-\!P_{\teta}(\ox^r)\|^2  \underbrace{-\frac{1}{2 \teta}\left\|{P_{\teta}(\ox^{r+1})}\!-\!\tilde{\bx}^{r+1}\right\|^2}_{(\rm IV )}\\  \nonumber
&+\underbrace{\left\langle {P_{\teta}(\ox^{r+1})}-\tilde{\bx}^{r+1}, \nabla f\left(P_{\teta}(\ox^r)\right)-\bv^r\right\rangle}_{(\rm V)} \big]. \nonumber
\end{align}
The term (IV) can be used to eliminate part of the term (V). In particular, since 
$\|a+b\|^2\le \frac{1}{2\teta}\|a\|^2+\frac{\teta}{2}\|b\|^2 $,
\begin{align}\label{eq:iv+v}
&({\rm IV})+({\rm V})\\ \nonumber
\leq&  ({\rm IV})+\frac{1}{2 \teta}\|{P_{\teta}(\ox^{r+1})}-\tilde{\bx}^{r+1}\|^2+\frac{{\teta}}{2}\|\nabla f(P_{\teta}(\ox^r))-\bv^r\|^2\\  \nonumber
=&\frac{{\teta}}{2}\|\nabla f(P_{\teta}(\ox^r))-\bv^r\|^2. 
\end{align}
Recall that $\nabla f(P_{\teta}(\ox^r))= \frac{1}{n} \sum_{i=1}^n \nabla f_i(P_{\teta}(\ox^r))$ $=\frac{1}{n\tau} \sum_{t=0}^{\tau-1} \sum_{i=1}^n\nabla f_i\left( P_{\teta}(\ox^r)\right)  $ and $\bv^r:=$ $\frac{1}{n\tau} \sum_{t=0}^{\tau-1}$ $ \sum_{i=1}^n\left( \nabla f_i\left(\bz_{i,t}^r;\mB_{i,t}^r\right)\right)$. By adding and subtracting $\nabla f_i(\bz_{i,t}^r)$ and following similar derivations as in \eqref{eq:corollary}, 
\begin{equation}\label{eq:v-g-26}
\begin{aligned}
& \bE \left\| \bv^r -\nabla f(P_{\teta}(\ox^r)) \right\|^2\\
=& \bE \big\| \frac{1}{n\tau} \sum_{t=0}^{\tau-1} \sum_{i=1}^n\Big( \nabla f_i\left(\bz_{i,t}^r;\mB_{i,t}^r\right)  -\nabla f_i(\bz_{i,t}^r)\\
&+\nabla f_i(\bz_{i,t}^r)   -\nabla f_i(P_{\teta}(\ox^r)) \Big) \big\|^2	\\
\le & 2L^2 \frac{1}{n\tau} \sum_{i=1}^{n} \sum_{t=0}^{\tau-1}{ \bE\|\bz_{i,t}^r-P_{\teta}(\ox^r)\|^2}  +\frac{2}{\tau n} \frac{\sigma^2 }{b},
\end{aligned}
\end{equation}
where the inequality follows from $L$-smoothness Assumption \ref{asm-smooth}.
Using \eqref{eq:iv+v} and \eqref{eq:v-g-26} in 
\eqref{eq:f-f-25} yields
\begin{align}\label{eq:x-bar-x*-24}
& \bE [F(P_{\teta}(\ox^{r+1}))]\\\nonumber
\le & \bE\Big[ F\left(P_{\teta}(\ox^r)\right)+\left(L-\frac{1}{2 \teta}\right)\left\|\tilde{\bx}^{r+1}-P_{\teta}(\ox^r)\right\|^2\\\nonumber
&+\left(\frac{L}{2}-\frac{1}{2 \teta}\right)\left\|{P_{\teta}(\ox^{r+1})}-P_{\teta}(\ox^r)\right\|^2\\\nonumber
&+\frac{\teta}{2}\left(  \frac{2L^2}{n\tau} \sum_{i=1}^{n} \sum_{t=0}^{\tau-1} {\|\bz_{i,t}^r-P_{\teta}(\ox^r)\|^2} +\frac{2}{\tau n} \frac{\sigma^2 }{b} \right)\Big]. 
\end{align}
The final term is the drift-error, which we bounded in Lemma~\ref{lem:phi-bar-x}. Using   \eqref{eq:phi--29}  in {\eqref{eq:x-bar-x*-24}} gives
\begin{align}\label{eq:y32}
&\bE [F(P_{\teta}(\ox^{r+1}))]\\\nonumber
\le& \bE \Big[F\left(P_{\teta}(\ox^r)\right)+\left(L-\frac{1}{2 \teta}\right)\left\|\tilde{\bx}^{r+1}-P_{\teta}(\ox^r)\right\|^2\\\nonumber
&+\left(\frac{L}{2}-\frac{1}{2 \teta}\right)\left\|{P_{\teta}(\ox^{r+1})}-P_{\teta}(\ox^r)\right\|^2 \\\nonumber
&+\frac{\teta}{2} \cdot2L^2  \frac{1}{n\tau}\cdot	1.15\tau \Big(
4\tau^2\eta^2 n\cdot 4 B_g^2  \\\nonumber
+& 4 n \tau^2\eta^2 \|\mathcal{G}_{\teta g}( P_{\teta}(\ox^r))  \|^2 + 4  \mathbb{E}\| \bLam^r \!-\!  \obLam^r \|^2+ 8n \tau \eta^2 \frac{\sigma^2}{b} \Big)\\\nonumber
&+\frac{\teta}{2}\cdot 2\frac{1}{\tau n} \frac{\sigma^2 }{b}\Big].
\end{align}

To quantify the  {auxiliary} function descent, we subtract $F^{\star}$ from both sides of \eqref{eq:y32} and multiply by $\tilde{\eta}$, then sum the resulting inequality with Lemma \ref{lem-Lambda} to obtain
\begin{align}\label{eqn:Lypnv}
\nonumber
&\bE \Big[ \teta \left(F(P_{\teta}(\ox^{r+1}))-F^{\star}\right)+ \frac{1}{n}\|\bLam^{r+1}-\obLam^{r+1}\|^2\Big] \\\nonumber
\le& \bE \big[\teta \left(F(P_{\teta}(\ox^{r})) -F^{\star}\right)+\frac{1}{n}\|\bLam^r\!-\!\obLam^r\|^2 + \frac{6\teta^2}{n\tau} \frac{\sigma^2}{b} \\
&+ \frac{ 2.8\cdot20 L^2\teta^4}{\eta_g^2} B_g^2-0.3\teta^2 \left\|\mathcal{G}(P_{\teta}(\ox^r))\right\|^2\big],
\end{align}
where the inequality follows after substituting the step-size condition \eqref{eq:step-sizes} and performing straightforward algebraic calculations.  
By the definition of the {auxiliary} function $\Omega^r$, \eqref{eqn:Lypnv} implies that
\begin{equation}\label{eq:xx-34}
\begin{aligned}
&\bE\left[\Omega^{r+1}\right]\le \bE\left[\Omega^{r}\right]  +  \frac{6\teta}{n\tau} \frac{\sigma^2}{b}\\
&+ \frac{2.8\cdot20L^2\teta^3}{\eta_g^2} B_g^2-{0.3\teta} \bE\left\|\mathcal{G}(P_{\teta}(\ox^r))\right\|^2.
\end{aligned}
\end{equation}
Telescoping this inequality completes the proof of Theorem \ref{thm:noncvx}.

\subsection{Proof of Theorem \ref{thm:pl}}\label{prf:pl}
The proximal PL inequality allows us to establish stronger function-value decrease by the virtual centralized PGD iterate \eqref{eq:pgd}. Following the procedure in \cite[Equation (34)]{j2016proximal}, we obtain 
\begin{equation}\label{eq:x36}
\begin{aligned}
&\mathbb{E}\left[F\left(\tilde{\bx}^{r+1}\right)\right]
\leq	
\mathbb{E}\left[F\left(P_{\teta}(\ox^{r})\right)-\mu \teta\left( F\left(P_{\teta}(\ox^{r})\right)-F^{\star}\right)\right],
\end{aligned}
\end{equation}
where we have used that  $\teta \leq 1/{(10L)}$.  Adding $2 / 3 \times$\eqref{eq:y22} and $1 / 3 \times$\eqref{eq:x36} yields 
\begin{align}\label{eq:pl-f-pgd}
&\mathbb{E}\left[F\left(\tilde{\bx}^{r+1}\right)\right] \leq \mathbb{E}\big[F\left(P_{\teta}(\ox^{r})\right)\\
&+\!\left(\frac{L}{3}-\frac{2}{3 \teta}\right)\left\|\tilde{\bx}^{r+1}-P_{\teta}(\ox^{r})\right\|^2\!-\!\frac{\mu \teta}{3}\left(F\left(P_{\teta}(\ox^{r})\right)\!-\!F^{\star}\right)\big]. \nonumber
\end{align}
Recall that \eqref{eq:y23} describes the function-value decrease of our algorithm at the server-side. Using \eqref{eq:pl-f-pgd} to eliminate the virtual function value $F(\tilde{\bx}^{r+1} )$ in \eqref{eq:y23} yields
\begin{align}\label{eq:pl-f-34}
&\mathbb{E}\left[F\left(P_{\teta}(\ox^{r+1})\right)\right] \\\nonumber
\le& \mathbb{E}\big[F\left(P_{\teta}(\ox^{r})\right)+\left(\frac{5 L}{6}-\frac{1}{6 \teta}\right)\left\|\tilde{\bx}^{r+1}-P_{\teta}(\ox^{r})\right\|^2\\\nonumber
&+\left(\frac{L}{2}-\frac{1}{2 \teta}\right)\left\|P_{\teta}(\ox^{r+1})-P_{\teta}(\ox^{r})\right\|^2 \\\nonumber
&-\frac{\mu \teta}{3}\left(F\left(P_{\teta}(\ox^{r})\right)-F^{\star}\right)\underbrace{-\frac{1}{2 \teta}\left\|P_{\teta}(\ox^{r+1})-\tilde{\bx}^{r+1}\right\|^2}_{\rm (IV)}\\\nonumber
&+\underbrace{\left\langle P_{\teta}(\ox^{r+1})-\tilde{\bx}^{r+1}, \nabla f\left(P_{\teta}(\ox^{r})\right)-\bv^r\right\rangle\big]}_{\rm (V)}.   
\end{align}
We encounter the same terms (IV)+(V) as in \eqref{eq:f-f-25} and proceed similarly. By using \eqref{eq:iv+v} and \eqref{eq:v-g-26} in 
\eqref{eq:pl-f-34} and then subtracting $F^{\star}$ from both sides, we find
\begin{align}\label{eq:xx-38}
& \mathbb{E}\left[F\left(P_{\teta}(\ox^{r+1})\right)-F^{\star}\right] \\\nonumber
\leq&\mathbb{E}\Big[ \left(1-\frac{\mu \teta}{3}\right) \left( F\left(P_{\teta}(\ox^{r})\right)\!-\! F^{\star}\right) \!-\! \frac{1}{12 \teta}\left\|\tilde{\bx}^{r+1}-P_{\teta}(\ox^r)\right\|^2\\\nonumber
& +\left(\frac{L}{2}-\frac{1}{2 \teta}\right)\left\|P_{\teta}(\ox^{r+1})-P_{\teta}(\ox^{r})\right\|^2 \\\nonumber
&+\frac{\teta}{2} \Big( \frac{2L^2}{n\tau} \sum_{i=1}^{n} \sum_{t=0}^{\tau-1} {\|\bz_{i,t}^r-P_{\teta}(\ox^r)\|^2}  +\frac{2}{\tau n} \frac{\sigma^2 }{b} \Big) \Big], 
\end{align}
where we have used that  $\frac{5 L}{6}-\frac{1}{6 \teta} \le \frac{1}{12\teta}$. Since~\eqref{eq:xx-38} and~\eqref{eq:x-bar-x*-24} only differ in the two first terms of the right-hand side,
%
we can follow similar arguments as we used in deriving~\eqref{eqn:Lypnv} and use Lemmas~\ref{lem:phi-bar-x} and~\ref{lem-Lambda} to find 
\begin{align}\label{eqn:Lypnv-PL}
\nonumber
&\bE\big[\teta (F(P_{\teta}(\ox^{r+1}))-F^{\star}) + \frac{1}{n}\|\bLam^{r+1}-\obLam^{r+1}\|^2\big] \\ \nonumber
\le&  \left(1-\frac{\mu \teta}{3}\right) \bE\Big[\teta \left(F(P_{\teta}(\ox^{r})) -F^{\star}\right)+\frac{1}{n}\|\bLam^r\!-\!\obLam^r\|^2\Big] \\
&+6\teta^2 \frac{1}{n\tau} \frac{\sigma^2}{b} + 2.8\cdot20 L^2 \frac{\teta^4}{\eta_g^2} B_g^2.
\end{align}
Also in this case, the final term is the result of the step-size condition \eqref{eq:step-sizes} and simple algebraic calculations.  By dividing both sides of the inequality with $\tilde{\eta}$ we obtain the following inequality for $\Omega^r$
%
\begin{equation*}
\begin{aligned}
\bE[\Omega^{r+1}]\le \left(1-\frac{\mu \teta}{3}\right) \bE[\Omega^{r}]+ \frac{6\teta}{n\tau} \frac{\sigma^2}{b} +  2.8\cdot20 L^2 \frac{\teta^3}{\eta_g^2} B_g^2. 
\end{aligned}
\end{equation*}
Telescoping this inequality yields the desired result and concludes the proof of Theorem \ref{thm:pl}.

\bibliographystyle{plain}    
\bibliography{autosam}  
\end{document}